\documentclass[]{artificer}
\usepackage{microtype}
\usepackage{amsfonts}
\usepackage{graphicx}
\usepackage{booktabs}
\usepackage{wrapfig}
\usepackage{amsmath}
\usepackage{amsthm}
\usepackage{marvosym}
\usepackage{algpseudocode}
\usepackage{algorithm}

\usepackage[capitalize,noabbrev]{cleveref}
\usepackage{enumitem}

\usepackage{tcolorbox}
\usepackage{xcolor}
\colorlet{tabcolor}{colorbg}
\def\ourname{\texttt{MotionEcho}}
\definecolor{skyblue}{RGB}{62, 176, 210}
\definecolor{inter_purple}{RGB}{153, 107, 187}
\definecolor{softpink}{RGB}{243, 210, 239}
\usepackage{pifont}
\usepackage{seqsplit}

\newcommand{\eg}{\textit{e.g.}}
\newcommand{\ie}{\textit{i.e.}}

\title{When Distillation Breaks Motion Control: Restoring Generative Trajectories for Fast Video Generators}

\author[1 \, 2 \, \spadesuit]{Jintao Rong}
\author[2 \, \spadesuit]{Xin Xie}
\author[1]{Xinyi Yu}
\author[1 \, {\textnormal{\Letter}}]{Linlin Ou}
\author[3]{Xinyu Zhang}
\author[1 \, 4]{Chunhua Shen}
\author[2 \, {\textnormal{\Letter}}]{Dong~Gong}
\affiliation[1]{Zhejiang University of Technology}
\affiliation[2]{University of New South Wales (UNSW Sydney)}
\affiliation[3]{University of Auckland}
\affiliation[4]{Zhejiang University}
\contribution{\textsuperscript{$\spadesuit$}Equal contribution.}
\contribution{\textsuperscript{\Letter}Corresponding authors.}
\contribution{The work was done during Jintao Rong's Research Practicum visit at UNSW Sydney.}

\abstract{
 Training-free motion customization imposes motion patterns from reference videos onto video generators through test-time computation.  Most existing methods target full diffusion models, requiring many denoising steps and high computational cost. With the rise of efficient distilled models, a natural question arises: \textit{can test-time motion customization be applied directly to distilled generators with their accelerated sampling and efficiency gains?} However, our analysis reveals that existing training-free techniques fail on distilled models. Distillation fundamentally alters the denoising dynamics that prior test-time guidance relies on, and the large denoising steps of distilled generators discard the dense intermediate states that score guidance requires, rendering existing motion control strategies incompatible with fast generation. To address this limitation, we propose \ourname, a novel training-free \emph{test-time distillation} framework that enables motion customization for distilled video generators. The key idea is to correct the student model's sampling trajectory with restricted usage of a high-quality diffusion teacher at inference time. Teacher supervises the student's denoising by re-noising the student's endpoint onto its dense trajectory to form a motion-aligned clean endpoint, then interpolating it with the student's, while an adaptive scheduling mechanism determines when and how much teacher guidance is needed. As a result, \ourname~restores generative trajectories for distilled video generators via lightweight, adaptive test-time teacher guidance, enabling accurate motion control without compromising generation efficiency. Extensive experiments on multiple distilled video generation models demonstrate that our method significantly improves motion fidelity and visual quality while retaining the efficiency advantages of distilled generation.
}

\metadata[Project Page]{\href{https://euminds.github.io/motionecho/}{https://euminds.github.io/motionecho/}}

\begin{document}

\maketitle

\section{Introduction}
\label{sec:intro}

Text-to-Video (T2V) diffusion models~\cite{VDM, Videocrafter1, GoogleVeo, OpenaiSora, T2vTurboV1, cogvideox, wan} have recently achieved remarkable progress in conditional video generation, enabling the synthesis of high-quality and temporally coherent videos from textual prompts. Beyond text guidance, a growing body of works has focused on motion customization~\cite{VMC, MotionDirector, MotionInversion, ControlVideo, zhao2023controlvideo, ControlAVideo, MotionClone, DMT, Text2VideoZero, MOFT, DitFlow, effivmt}, which aims to generate videos that follow motion patterns extracted from reference videos while synthesizing novel visual content from text descriptions.

Existing motion customization approaches can be broadly categorized into training-based and training-free methods. Training-based methods incorporate motion information into T2V models via LoRA adapter~\cite{MotionDirector}, temporal attention adaptation \cite{VMC}, motion embeddings~\cite{MotionInversion}, and ControlNet-style condition branches~\cite{ControlVideo,ControlAVideo,zhao2023controlvideo}.
While effective, such approaches require costly optimization for each motion condition. To enable more flexible motion control, training-free methods~\cite{MotionClone, DMT, Text2VideoZero, MOFT, DitFlow} inject motion patterns via test-time computation through optimization or regularization-based guidance, making them adaptable across different models and motion inputs. However, despite their flexibility, most existing training-free motion customization methods are designed for full diffusion models, whose iterative sampling process involves dozens of denoising steps. The training-free methods introduce additional computations on top of it.

\begin{figure}[t]
    \centering
    \includegraphics[width=1.0 \linewidth]{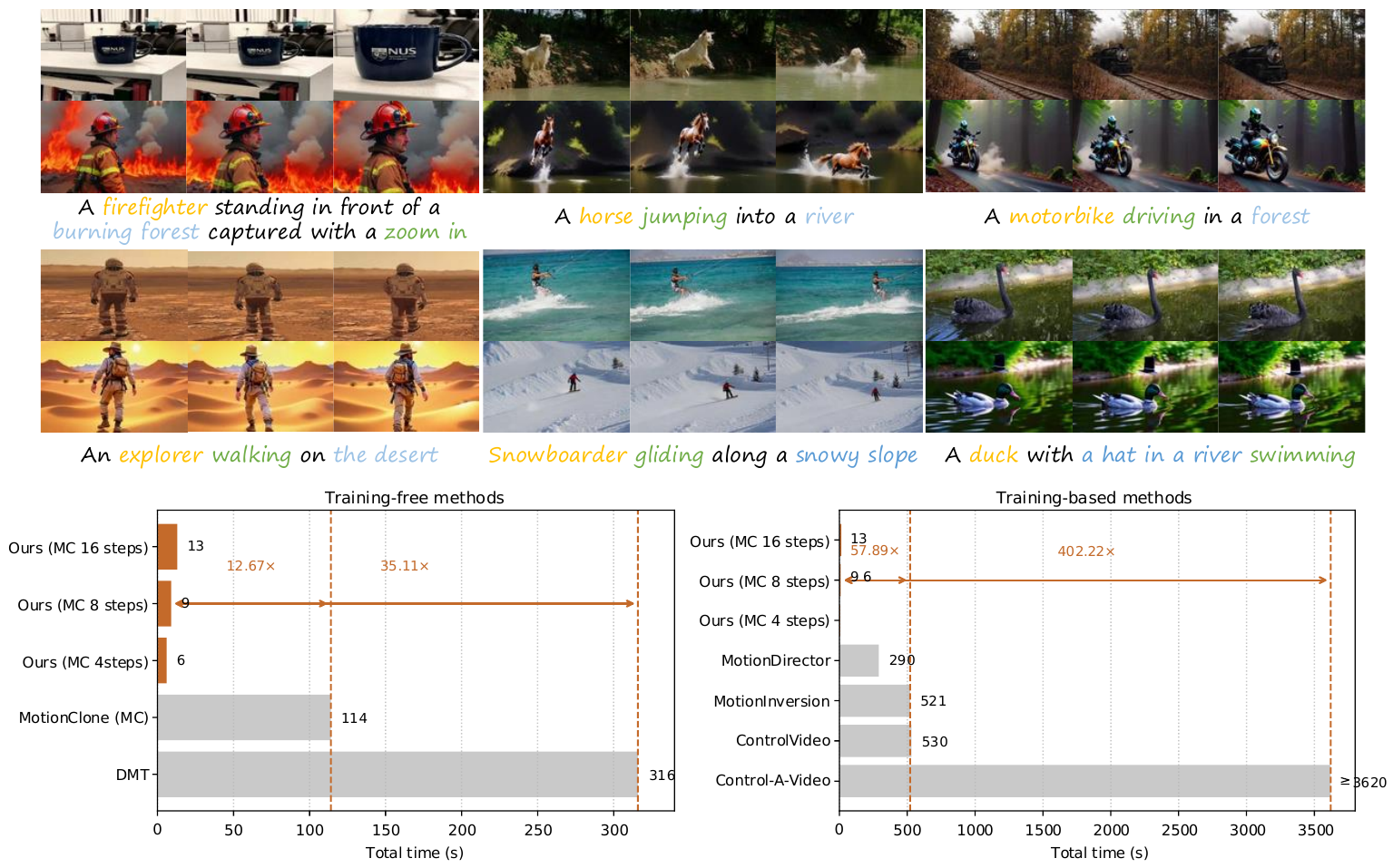}
    \caption{
    \ourname~is a training-free motion customization framework that enables effective motion control on distilled video generators through adaptive, teacher-guided test-time distillation. It achieves high-quality motion transfer across diverse motion types (top) while preserving the efficiency advantages of distilled generation (bottom).}
    \label{fig:teaser}
\end{figure}

\begin{figure}[t]
    \centering
    \includegraphics[width=1.0\linewidth]{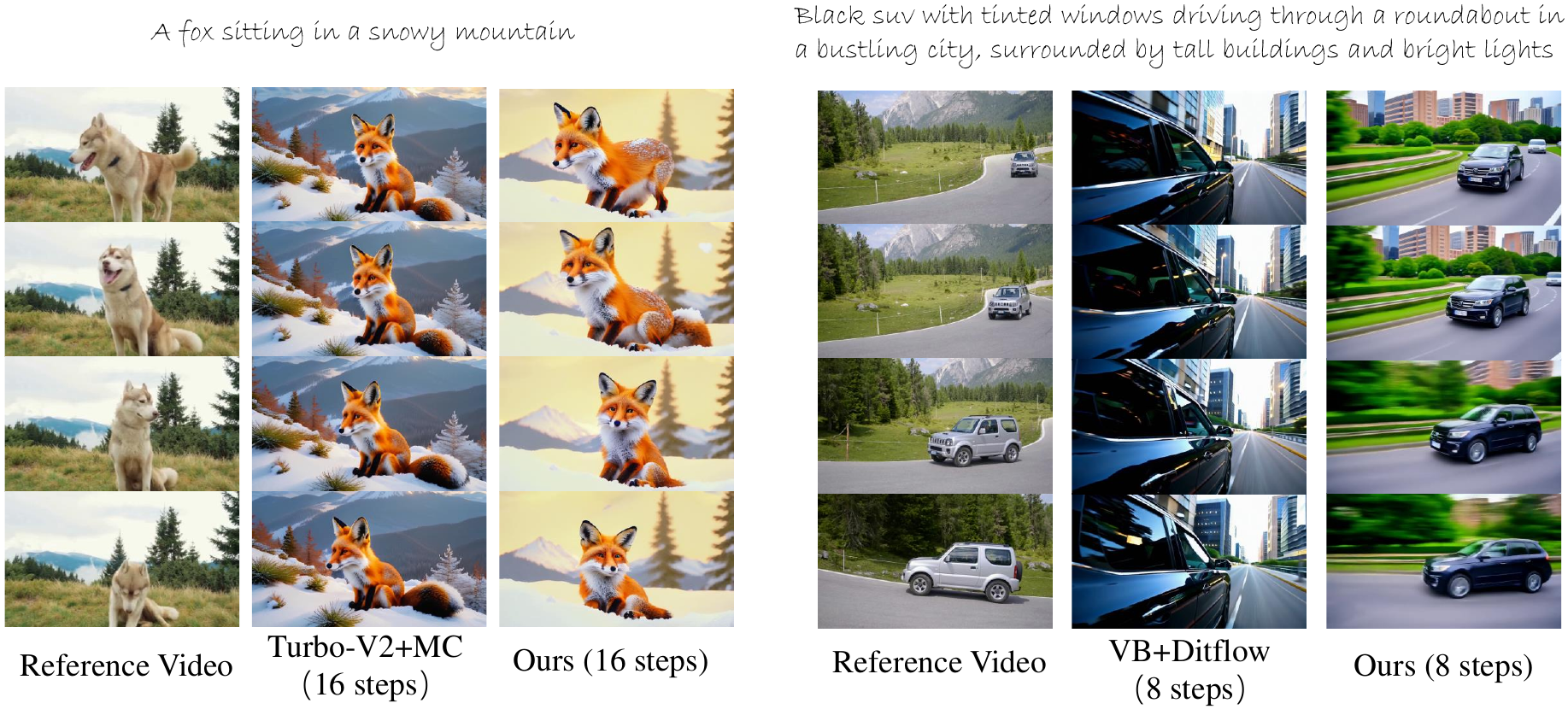}
    \caption{
    \textbf{Visual comparison of motion customization on fast distilled video models.} Directly applying existing training-free methods to distilled models leads to severe degradation. For instance, the baseline fails to mimic the reference object motion (left, the fox fails to raise its head) and suffers from catastrophic structural degradation under dynamic camera motion (right, the background and car shape become corrupted). In contrast, our proposed \ourname~successfully transfers the reference motion while maintaining high visual quality and strict text alignment during a highly accelerated few-step inference regime.}
    \label{fig:comparison}
\end{figure}

With the development of distilled T2V generators~\cite{videolcm,T2vTurboV1,T2vTurboV2,AnimateDiffLightning,TDM,Videoblade}, which compress the multi-step denoising process of a teacher diffusion model into a few-step student model via distillation, inference is significantly accelerated while maintaining competitive generation quality. \textit{This motivates a key question: can training-free test-time motion customization be adapted for distilled T2V models with altered generative trajectories? } To investigate, we conducted preliminary experiments by directly applying representative training-free motion customization methods, including MotionClone~\cite{MotionClone} and DiTFlow~\cite{DitFlow}, to distilled T2V models including T2V-Turbo-V2~\cite{T2vTurboV2} and Video-Blade~\cite{Videoblade}. Fig.~\ref{fig:comparison} shows the generated videos exhibit significant motion degradation, \ie, the motion cannot be exactly transferred from reference videos to targeted ones. We attribute these failures to fundamental differences between standard diffusion models and distilled generators: i) distilled models collapse multiple denoising steps into a small number of large-step updates, making the progressive interventions used in existing training-free methods \textit{overly coarse and ineffective}; ii) the denoising dynamics of distilled models, trained to \textit{approximate multi-step trajectories}, differ substantially from those of the teacher models with \textit{dense} timesteps, rendering prior motion guidance strategies incompatible. The large denoising steps of distilled generators discard the dense intermediate states that score guidance requires.

To address this, we propose \textbf{\ourname}, a novel training-free framework that restores generative trajectories for distilled T2V models to enable effective test-time motion customization, while preserving the efficiency of the distilled fast generator. The key idea is to correct the student model's disrupted sampling trajectory using selective teacher guidance during inference, as a form of complementary test-time distillation, which bridges the dynamic gap without replacing the efficient distilled backbone. 
Specifically, the distilled model serves as the primary inference backbone to maintain efficient generation, while a high-quality teacher (full) diffusion model provides auxiliary supervision during selected reverse timesteps. Through endpoint prediction and interpolation, the teacher guidance acts as an \emph{echo} that corrects the student’s sampling trajectory and enables accurate motion transfer. 
To preserve efficiency, we further introduce an \textit{adaptive acceleration strategy} that dynamically determines when teacher supervision is necessary and decides how to adjust the optimization budget at each denoising step. This adaptive scheduling balances motion fidelity and computational efficiency during inference.
Quantitative and qualitative experiments demonstrate that our \ourname~enables motion customization success on distilled models. As illustrated in \cref{fig:teaser}, 
different motion patterns extracted from reference videos can be accurately transferred to targeted generated videos, while keeping high inference speed.
Our contributions are as follows:

\begin{itemize}
    \item We first identify the limitations of applying existing training-free motion customization methods to fast distilled T2V models and provide empirical and analytical evidence of the resulting motion degradation.
    
    \item We propose \ourname, a novel training-free test-time distillation framework that enables motion customization for distilled video diffusion models via teacher-guided test-time distillation during inference.
    
    \item We introduce an adaptive acceleration strategy that dynamically schedules when and how much teacher supervision is applied at each denoising step, effectively improving motion fidelity without sacrificing efficiency.
    
    \item We demonstrate through extensive experiments on multiple distilled video diffusion models that \ourname \\ achieves superior motion fidelity, visual quality, and efficiency compared to other methods across two benchmarks.
\end{itemize}

\section{Related Work}
\label{Related_Work}
\subsection{Text-to-Video Diffusion Models}
With the advent of diffusion models~\cite{DDIM, DDPM, SDE}, remarkable advancements in content creation have been witnessed in the text-to-image field~\cite{GLIDE, FLUX, SD3, IterComp, SD, DALLE3, imagen,SDXL, DyMO,  HyperAlign,lin2026narratology,jha2025mining}. Following this success, recent text-to-video diffusion models, such as Sora~\cite{OpenaiSora}, Gen-3 Alpha~\cite{RunwayGEN3}, and Veo~\cite{GoogleVeo}, have achieved high-quality video generation but remain unavailable to the public.
To break this limitation, the seminal work VDM~\cite{VDM} leverages a 3D UNet~\cite{3DUnet}, factorized over space and time, for unconditional video generation. Subsequently, Imagen Video~\cite{ImagenVideo} and Make-A-Video~\cite{MakeAVideo} adopt cascade frameworks to enable high-resolution text-conditioned video generation in pixel space. To reduce the computational cost, LVDM~\cite{LVDM}, VideoLDM~\cite{VideoLDM}, and MagicVideo~\cite{MagicVideo} extend diffusion models to a 3D latent space, facilitating training on large-scale video datasets~\cite{WebVID, HD_VILA_100M}. 

Rather than training from scratch, AnimateDiff~\cite{AnimateDiff} trains a domain adapter for quality enhancement and a temporal module for motion prior learning via lightweight LoRAs~\cite{LoRA}. 
VideoCrafter2~\cite{Videocrafter2} constructs a UNet with temporal attention layers and uses low-quality videos for motion learning while using high-quality images for appearance learning.
CogVideoX~\cite{cogvideox} uses a Diffusion Transformer (DiT) with expert transformers and multiple resolution training for coherent videos.
Wan~\cite{wan} advances large-scale video generation by using a diffusion transformer framework.
T2V-Turbo~\cite{T2vTurboV1} and T2V-Turbo-V2~\cite{T2vTurboV2} adopt consistency models~\cite{ConsistencyModels} to distill from VideoCrafter2~\cite{Videocrafter2} for accelerating the sampling process. 
Video-Blade~\cite{Videoblade} jointly trains adaptive block-sparse attention with sparsity-aware step distillation, enabling data-free distillation. OPAD~\cite{OPAD} explores training-based concept personalization for one-step generation.

\subsection{Video Motion Customization}
Motion customization aims to enable the generated videos to follow specific motion patterns from reference videos while remaining semantically aligned with textual descriptions~\cite{TuneAVideo,zhao2023controlvideo,ControlAVideo,VMC,MotionDirector,li2026let,MotionInversion,MotionClone,MOFT,PointMotion,liao2024evaluation,DitFlow,VideoMSG}. The pioneering work Tune-A-Video~\cite{TuneAVideo} introduces one-shot video generation by tuning SD~\cite{SD} on a single reference video. ControlVideo~\cite{zhao2023controlvideo} and Text2Video-Zero~\cite{Text2VideoZero} inherit priors from ControlNet~\cite{ControlNet} and leverage cross-frame interactions to enable zero-shot controllable video synthesis. Furthermore, Control-A-Video~\cite{ControlAVideo} incorporates trainable motion layers into UNet and ControlNet~\cite{ControlNet} for temporal modeling, achieving motion control conditioned on depth, sketch, or motion information from reference videos. Despite achieving high-quality customization, these methods still suffer from motion misalignment. 

To address this, existing motion-specific tuning methods, such as VMC~\cite{VMC}, MotionDirector~\cite{MotionDirector}, FlexiMMT~\cite{li2026let}, and MotionInversion~\cite{MotionInversion}, are designed to decouple the learning of appearance and motion, enabling the generalization of distinct motion patterns to diverse textual prompts and scenes. Additionally, DMT~\cite{DMT} and MotionClone~\cite{MotionClone} employ training-free strategies, extracting motion priors from the latent representations of reference videos and constructing an energy function to guide motion customization during inference.
MOFT~\cite{MOFT} is a training-free approach that uses Principal Component Analysis (PCA)~\cite{pca} to derive motion subspaces from the video diffusion model to control video motion.
Zhang \emph{et al.}~\cite{PointMotion} propose a motion consistency loss combined with inverted reference noise to enhance temporal coherence and motion accuracy.
DiTFlow~\cite{DitFlow} transfers motion in DiT-based T2V models by extracting Attention Motion Flow (AMF) from cross-frame attention and performing latent optimization for motion guidance. FreeNoise~\cite{FreeNoise} and FreeTraj~\cite{FreeTraj} control video motion by rescheduling the initial noise. Furthermore, Video-MSG~\cite{VideoMSG} generates a video motion sketch through multimodal planning and injects it into T2V generators via noise inversion.


\section{Methods}
\begin{figure*}[t]
    \centering
    \includegraphics[width=1.0\linewidth]{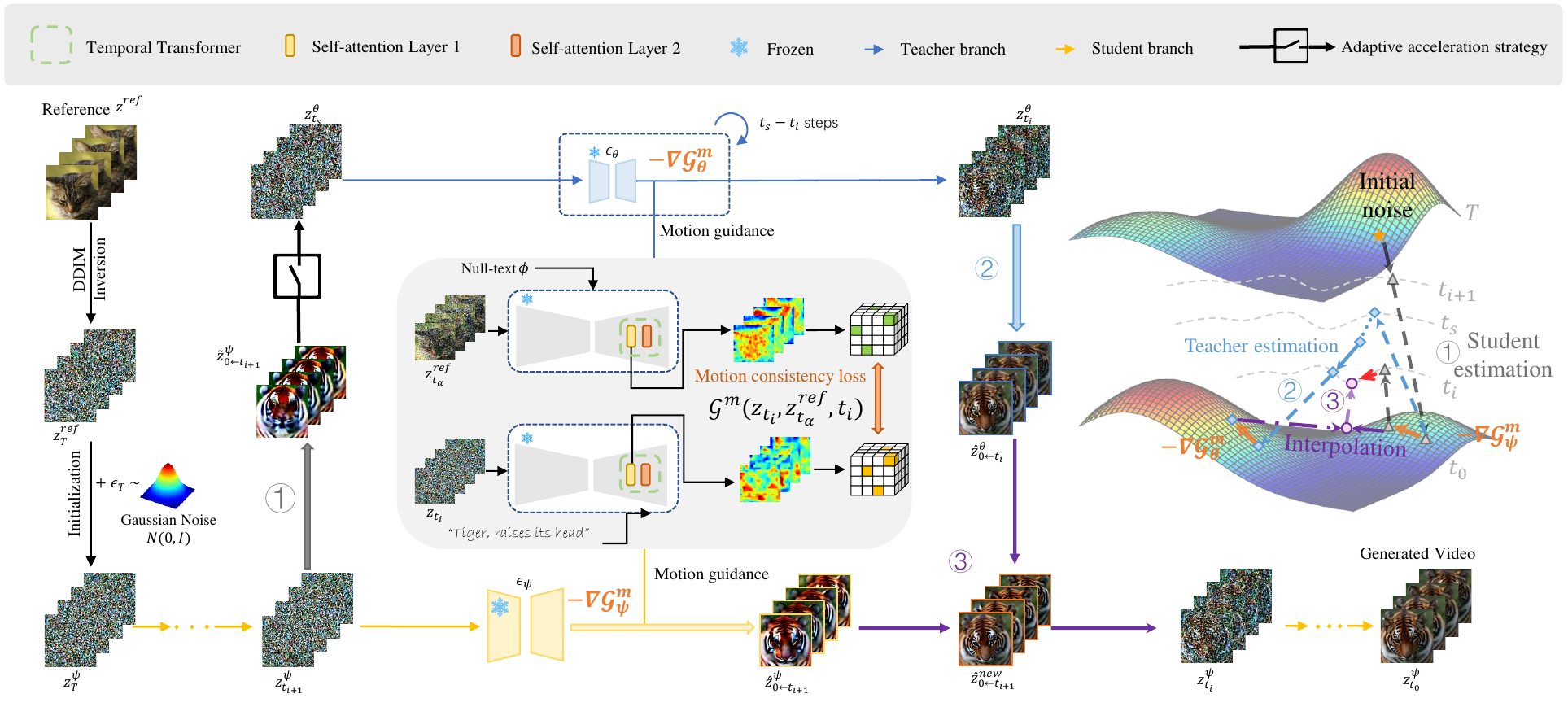}
    \caption{\textbf{Pipeline of MotionEcho.} Given a reference video, motion priors are extracted to initialize the student model with a motion-preserving noisy latent. During inference, the teacher (\textcolor{skyblue}{blue path \ding{173}}) and student (\textcolor{gray}{gray path \ding{172}}) models perform motion customization using motion loss gradients. Teacher guidance is applied via prediction interpolation (\textcolor{inter_purple}{purple path \ding{174}}) at sub-interval endpoints. The student then generates the final video in a few steps with high motion fidelity. }
    \label{fig:pipeline}
\end{figure*}
We first establish a diagnostic baseline by directly applying existing training-free motion customization methods to distilled models, observing significant visual artifacts and temporal inconsistencies arising from coarse denoising and mismatched sampling dynamics. To address this, we propose \ourname, a training-free framework that restores generative trajectories to enable effective motion customization in fast video generators.
By strictly maintaining the distilled student as the primary inference backbone, \ourname~leverages a fine-grained teacher diffusion model as an on-demand trajectory corrector via score distillation. To maintain efficiency, we further develop an adaptive acceleration strategy that triggers the teacher's guidance and truncates its denoising iterations.
The whole pipeline of \ourname~is demonstrated in Fig.~\ref{fig:pipeline}.

\subsection{Preliminaries}
\paragraph{Video Diffusion Models and Motion Customization.} 
Video diffusion models~\cite{DALLE3,AnimateDiff, VideoLDM, Videocrafter2, cogvideox, wan} generate videos by progressively denoising random noise through a learned reverse diffusion process. 
Given an input video $x$, a pre-trained encoder $\mathcal{E}$ maps it into the latent space $z_{0} = \mathcal{E}(x)$, which is gradually corrupted into $z_t= \sqrt{\bar \alpha_{t}}z_0 + \sqrt{1-\bar \alpha_{t}} \epsilon_{t}$, where $\epsilon_t \sim \mathcal{N}(0, \mathbf{I})$ represents Gaussian noise and $\bar \alpha_{t}$ is a hyperparameter for controlling the diffusion process.
To capture the data distribution, a neural network $\epsilon_\theta$ is trained to estimate the added noise from the noisy input $z_t$, which is formulated as follows:
\begin{equation}
    \min_{\theta} \mathbb{E}_{\mathcal{E}(x),\, \epsilon_{t} \sim \mathcal{N}(0,\mathbf{I}),\, t \sim \mathcal{U}(1, T)} \left[ \left\| \epsilon_{t} - \epsilon_\theta(z_t, c, t) \right\|_2^2 \right],
\end{equation}
where $c$ is the condition such as the textual prompt. During inference, the sampling process begins with standard Gaussian noise. 
To enhance controllability, we typically introduce classifier-free guidance~\cite{CFG} or an additional customized energy function $\mathcal{G}(z_t, t)$~\cite{FreeDoM,TFG,MotionClone,DMT,MOFT} during sampling:
\begin{equation}
    \hat{\epsilon}_\theta = \epsilon_\theta(z_t, c, t) + \omega \big(\epsilon_\theta(z_t, c, t) - \epsilon_\theta(z_t, \phi, t)\big) - \eta \nabla_{z_t} \mathcal{G}(z_t, t),
\end{equation}
where $\phi$ denotes the unconditional null prompt, while $\omega$ and $\eta$ are scaling factors that determine the strength of classifier-free guidance and additional guidance, respectively.
From this viewpoint, motion customization can be instantiated by defining $\mathcal{G}$ as a motion energy that aligns the generated video with a reference motion pattern.
While training-based methods~\cite{MotionDirector, MotionInversion, TuneAVideo, ControlAVideo, ControlVideo, zhao2023controlvideo} adapt the model via fine-tuning or LoRA~\cite{LoRA},
training-free approaches~\cite{DMT, MotionClone, PointMotion, MOFT, DitFlow} inject motion priors at test time by optimizing the sampling trajectory through motion-conditioned guidance.
Both paradigms target the same objective of minimizing the mismatch between motion representations extracted from noisy latents,
$\mathbb{E}_{z_0, z^{\text{ref}}, t, \epsilon_t}\!\left[\left\|\mathcal{M}(z_t^{\text{ref}})-\mathcal{M}(z_t^{g})\right\|_2^2\right]$,
where $\mathcal{M}(\cdot)$ is a motion feature extractor, $z_t^{\text{ref}}$ is the noisy reference latent, and $z_t^{g}$ is the generated latent.

\paragraph{Distillation for Accelerated Video Diffusion Models.} 
The inherently iterative nature of video diffusion models results in slow inference and substantial computational overhead. 
To alleviate this, recent efforts leverage knowledge distillation~\cite{AnimateDiffLightning, AnimateLCM, T2vTurboV1, T2vTurboV2, TDM, Videoblade} to compress generation process. Our framework is compatible with two primary distillation paradigms. For UNet-based models (e.g., T2V-Turbo-V2~\cite{T2vTurboV2}), methods typically adopt consistency distillation~\cite{ConsistencyModels}. A student model $\epsilon_{\psi}$ is trained to approximate the macroscopic denoising trajectories of the teacher $\epsilon_{\theta}$ under a coarser timestep schedule $\mathcal{U}_\psi(1, T)$, formulated as $\min_{\psi} \, \mathbb{E}_{z_0, c, t \sim \mathcal{U}_\psi} \left[ \left\| \epsilon_{\psi}(z_{t}, c, t) - \epsilon_{\theta}(z_{t}, c, t) \right\|^2_2 \right]$.
Conversely, for DiT-based architectures (e.g., Video-Blade~\cite{Videoblade}), the generation process is accelerated via trajectory distribution distillation~\cite{TDM}. The student model $\epsilon_{\psi}$ is trained to match, in a single large step, the result of multiple teacher steps over an interval $[t_{i-1}, t_i]$. Formally, the objective is $\min_{\psi} \, \mathbb{E}_{z_{t_i}, c, i} \left[ \left\| \epsilon_{\psi}(z_{t_i}, c, t_i) - \tilde{\epsilon}_{\theta}(z_{t_i}, c, t_i \!\to\! t_{i-1}) \right\|^2_2 \right]$, where $\tilde{\epsilon}_{\theta}$ is the equivalent one-step noise prediction derived from the teacher's multi-step ODE solver output $\Phi_{\theta}(z_{t_i}, t_i \!\to\! t_{i-1})$.

\begin{figure*}[t]
    \centering
    \includegraphics[width=1.0\linewidth]{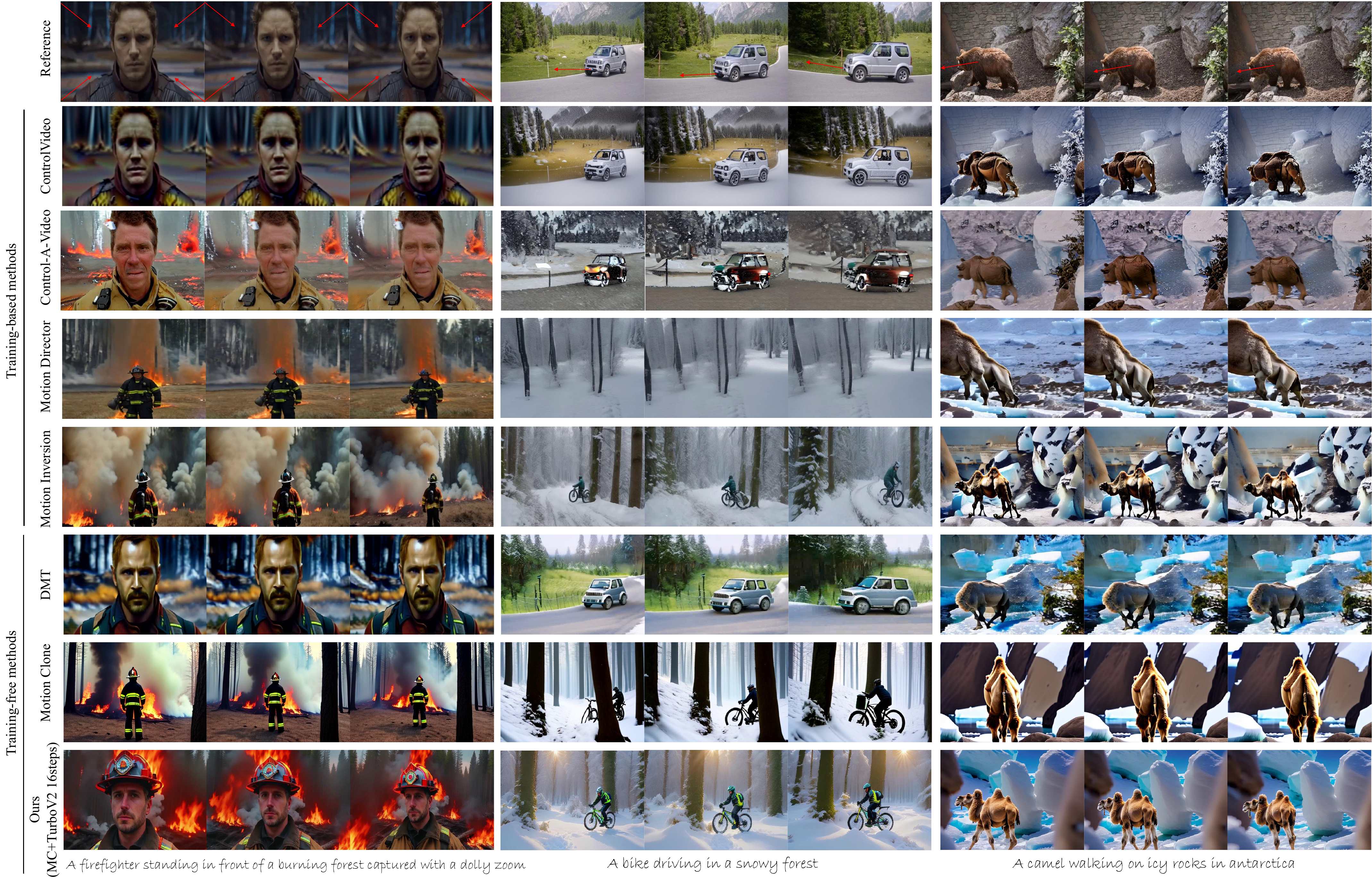}
    \caption{\textbf{Qualitative comparisons on VideoCrafter2-based methods.} Our method enables unified object, camera, and hybrid motion transfer with high motion fidelity and low inference time.}
    \label{fig:compare1}
\end{figure*}

\begin{figure*}[t]
    \centering
    \includegraphics[width=1.0\linewidth]{fig/benchmark_wan_motionguidance.pdf}
    \caption{\textbf{Qualitative comparisons on Wan2.1-1.3B-based methods.} 
    Our method achieves consistent high-fidelity motion preservation across diverse scenarios while maintaining faster inference than baseline methods.
    }
    \label{fig:compare2}
\end{figure*}

\subsection{Test-time Motion Customization for Fast Video Generators}
\label{subsec:test_time_student}
To enable training-free motion customization on distilled, fast T2V models and preserving efficiency, we explore an intuitive solution that directly integrates motion control~\cite{MotionClone,MOFT,DMT,DitFlow} into the sampling process of the distilled video diffusion model $\epsilon_{\psi}$. 
This solution retains the efficiency of the distilled student model while injecting motion representations without any training. 
For each large denoising step from $t_{i+1}$ to $t_i$ in the distilled model, the predicted noise is modified as: 
\begin{equation}
\begin{split}
\hat{\epsilon}_{\psi}(z^{\psi}_{t_{i+1}}, c, t_{i+1}) &= \tilde{\epsilon}_{\psi}(z^{\psi}_{t_{i+1}}, c, t_{i+1}) - \eta \nabla_{z^{\psi}_{t_{i+1}}} \mathcal{G}^m(z^{\psi}_{t_{i+1}}, z_{t_{\alpha}}^{\text{ref}}, t_{i+1})
\end{split}
\label{eq:modified_noise}
\end{equation}
where $\tilde{\epsilon}_{\psi}(z^{\psi}_{t_{i+1}}, c, t_{i+1}) = (1+\omega_{^{\psi}})\epsilon_{\psi}(z^{\psi}_{t_{i+1}}, c, t_{i+1}) - \omega_{^{\psi}} \epsilon_{\psi}(z^{\psi}_{t_{i+1}}, \phi, t_{i+1})$ denotes classifier-free guidance~\cite{CFG}. 
The proposed \ourname~is compatible with diverse inference-time guidance strategies and generative models. 
As a representative case, we use MotionClone~\cite{MotionClone} on diffusion models to illustrate the design of \ourname. Extensions to other guidance types and to flow-based models are provided in Section~\ref{ditmodel} and the supplementary material. 

Following MotionClone~\cite{MotionClone}, $\mathcal{G}^{m}(z^{\psi}_{t_{i+1}}, z^{\text{ref}}_{t_{\alpha}}, t_{i+1}) = || M^{\text{ref}}_{t_{\alpha}} \cdot ( \mathcal{A}(z^{\text{ref}}_{t_{\alpha}}) - \mathcal{A}(z^{\psi}_{t_{i+1}})) ||_2^2$ is the motion loss function, where $\mathcal{A}(\cdot)$ represents the temporal attention map and ${M}_{t_{\alpha}}^{\text{ref}}$ is a temporal mask derived from $\mathcal{A}(z^{\text{ref}}_{t_{\alpha}})$ at a fixed timestep $t_{\alpha}$ to enforce focused motion supervision. 
We only apply motion guidance during the initial $\tau$ percent of the sampling process. 
Then, $z^{\psi}_{t_{i}}$ is obtained by re-noising the predicted clean latent:
\begin{equation}
    z^{\psi}_{t_i} = \sqrt{\bar{\alpha}_{t_i}} \hat{z}^\psi_{0\leftarrow t_{i+1}} + \sqrt{1-\bar{\alpha}_{t_i}} \epsilon_{t_{i}},
\label{eq:renoise_t}
\end{equation}
where $\hat{z}^\psi_{0\leftarrow t_{i+1}} = \frac{z^{\psi}_{t_{i+1}}-\sqrt{1-\bar{\alpha}_{t_{i+1}}}\hat{\epsilon}_{\psi}(z^{\psi}_{t_{i+1}}, c, t_{i+1})}{\sqrt{\bar{\alpha}_{t_{i+1}}}}$ is the one-step prediction result. 
The distilled model generates results in a few denoising steps, during which motion customization is applied. 

However, empirical evidence reveals its fundamental inadequacy for motion customization quality and effectiveness. 
As shown in Fig.~\ref{fig:comparison}, the generated videos exhibit significant motion degradation, particularly in scenarios with complex object layouts or rapid movements. 
The degradation is further corroborated quantitatively in Table~\ref{tab:ablation}.
The cause of these failures lies in the distinct generative dynamics of distilled models. Specifically, their sparse timestep schedule forces motion gradients to act across much larger denoising intervals. This loss of control granularity transforms what should be progressive guidance into abrupt and unstable drifts in the latent space. Furthermore, distillation changes the sampling paradigm. While teacher models refine latents through a dense, incremental path, student models are trained to take discrete leaps toward macroscopic endpoints. Injecting localized motion priors into this highly compressed, large-step generation process creates a severe dynamic mismatch, rendering existing motion guidance mechanisms ineffective.

\subsection{Restoring Trajectories via Teacher-Guided Distillation}

\paragraph{Teacher-Guided Trajectory Correction.}

To address these limitations, we introduce a selective test-time distillation framework that uses a fine-grained teacher to correct the student.
We treat the student’s predicted clean latent with motion guidance, $\tilde{z}^\psi_{0\leftarrow t_{i+1}} = \frac{z^{\psi}_{t_{i+1}}-\sqrt{1-\bar{\alpha}_{t_{i+1}}}\tilde{\epsilon}_{\psi}(z^{\psi}_{t_{i+1}}, c, t_{i+1})}{\sqrt{\bar{\alpha}_{t_{i+1}}}}$, as an anchor.
To bridge the student’s large step with the teacher’s dense schedule, we renoise this anchor to an intermediate timestep $t_s \in [t_{i}, t_{i+1}]$ on the teacher’s trajectory.
Then, the teacher model follows Eq.~(\ref{eq:modified_noise}) and Eq.~(\ref{eq:renoise_t}) to perform iterative motion customization: $z_{t_{s-1}}^{\theta} \leftarrow z_{t_{s}}^{\theta} + \tilde{\epsilon}_\theta(z_{t_{s}}^{\theta}, c, t_{s}) - \eta \nabla_{z_{t_{s}}^{\theta}} \mathcal{G}^{m}(z_{t_{s}}^{\theta}, z_{t_\alpha}^\text{ref}, t_{s})$. This multi-step process yields a highly accurate, motion-aligned prediction $\hat{z}^{\theta}_{0\leftarrow t_{i}}$.

\paragraph{Score Interpolation as an Echo.}
To preserve the efficiency of the distilled model, we formulate the trajectory correction as a decomposed optimization problem. 
To transfer the teacher's corrected trajectory back to the fast sampling process, we define a score distillation sampling (SDS) objective that aims to align the student's estimate with the teacher's refined prediction: $\ell_{\text{distill}}= || \tilde{z}_{0 \leftarrow t_{i+1}}^{\psi} - \text{sg}\left[\hat{z}_{0 \leftarrow t_{i}}^{\theta}\right] ||_2^2$. 
During inference, directly minimizing this objective w.r.t. the intermediate noisy state $z^{\psi}_{t_{i+1}}$ requires computing the Jacobian $\frac{\partial \tilde{\epsilon}_\psi}{\partial z^{\psi}_{t_{i+1}}}$. This requires backpropagating through the entire network, which is fundamentally intractable. To solve this bottleneck, we decouple the prediction from the network by treating the student's predicted clean latent $\tilde{z}_{0 \leftarrow t_{i+1}}^{\psi}$ 
as an independent optimization variable.  
This formulation shifts the gradient update from the complex, high-curvature noisy score space directly to the clean data manifold. By evaluating the gradient step exclusively on the denoised estimate, the network Jacobian is completely bypassed. With the target fixed, the analytical gradient is straightforwardly $\nabla_{\tilde{z}_{0 \leftarrow t_{i+1}}^{\psi}} \ell_{\text{distill}} = 2(\tilde{z}_{0 \leftarrow t_{i+1}}^{\psi} - \hat{z}_{0 \leftarrow t_{i}}^{\theta})$. The gradient descent step elegantly reduces to a closed-form linear interpolation:
\begin{equation}
\begin{aligned}
\hat{z}_{0\leftarrow t_{i+1}}^{\text{new}} &= \tilde{z}^\psi_{0\leftarrow t_{i+1}} - \frac{\lambda}{2} \nabla_{\tilde{z}_{0 \leftarrow t_{i+1}}^{\psi}} \ell_{\text{distill}} = (1 - \lambda) \tilde{z}^{\psi}_{0\leftarrow t_{i+1}} + \lambda \hat{z}^{\theta}_{0\leftarrow t_{i}},
\end{aligned}
\label{eq:gradient_to_interpolation}
\end{equation}
where $\lambda$ controls the step size of the gradient descent, effectively acting as the strength of the teacher's restorative echo, and $\hat{z}_{0\leftarrow t_{i+1}}^{\text{new}}$ is the corrected student's predicted endpoint. Finally, the student model proceeds to the next noisy manifold $t_i$ using this interpolated clean latent:
\begin{equation}
z^{\psi}_{t_i} = \sqrt{\bar{\alpha}_{t_i}} \hat{z}_{0\leftarrow t_{i+1}}^{\text{new}} + \sqrt{1-\bar{\alpha}_{t_i}} \epsilon_{t_{i}}.
\label{eq:renoise_t_new}
\end{equation}

\subsection{Adaptive Acceleration Strategy for Efficiency}

Applying uniform teacher supervision across the initial $\tau$ percent of sampling is both computationally prohibitive and counterproductive due to varying denoising dynamics.
To maintain efficiency, we introduce an adaptive acceleration strategy comprising two core mechanisms. i) \emph{Step-wise Guidance Activation.} The decision to apply teacher guidance at each step becomes critical, which significantly affects the effectiveness of motion transfer. To make this decision more robust, we evaluate the necessity of distillation by computing the average motion loss over a moving window of previous steps, ${\mathcal{G}}^{\psi}_{t_i} = \frac{1}{W} \sum_{j=0}^{W-1} \mathcal{G}^m(z^{\psi}_{t_{i+j}}, z_{t_\alpha}^{\text{ref}}, t_{i+j})$, where $t_i$ denotes the current timestep and $W$ is the window size. If the average motion loss ${\mathcal{G}}^{\psi}_{t_i}$ exceeds a predefined threshold $\delta_1$, we activate teacher guidance for this step.
ii) \emph{Dynamic Truncation.} Once guidance is triggered, we control the internal guidance process of the teacher model via a dynamic truncation based on the smoothness of the motion loss. 
Specifically, during the interval from $t_s$ to $t_{i}$, we monitor the motion loss $\mathcal{G}^m(z^{\theta}_{t_{s}}, z_{t_\alpha}^{\text{ref}}, t_{s})$. If the loss falls below a confidence threshold $\delta_2$, we terminate the motion transfer early and directly denoise the current latent to the latent of $t_{i}$.

\subsection{MotionEcho for DiT-based Video Generators} 
\label{ditmodel}
Since adaptive distillation operates at the trajectory level, \ourname~accommodates different backbones and motion extractors $\mathcal{M}(\cdot)$.
DiT-based video generation methods achieve superior performance by jointly processing spatiotemporal information through attention mechanisms. 
Our method integrates both distilled DiT student models and teacher models with two key adaptations.
Unlike UNet-based methods (\eg, MotionClone~\cite{MotionClone}) that assume separable temporal attention, DiTs employ full spatiotemporal attention where motion and content are entangled. Therefore, we adopt generalized motion extractors (MOFT~\cite{MOFT}, DMT~\cite{DMT}) and DiT-specific methods (DiTFlow~\cite{DitFlow}). 
Once reference motion representations are extracted from DiT, we follow the test-time distillation paradigm with teacher motion guidance at specific steps to optimize $z^{\psi}_{t_i}$. Following DiTFlow~\cite{DitFlow}, we inject motion priors by processing $z_0^{\text{ref}}$ through the DiT and injecting the resulting keys $K$ and values $V$ into the target attention mechanism.

\section{Experiments}
\label{Experiments}

We evaluate \ourname~on a wide range of distilled video generators, comparing against state-of-the-art training-based and training-free motion customization methods. We further analyze each component's contribution via ablation studies.

\subsection{Experimental Setting}
\paragraph{Implementation Details.} 
Without requiring additional training, our method can be seamlessly integrated with various generative video diffusion models. To facilitate discussion, we denote the models by their abbreviations in parentheses. During inference, VideoCrafter2~\cite{Videocrafter2} (VC2) is employed as the teacher model to provide guidance for the distilled model, T2V-Turbo-V2~\cite{T2vTurboV2} (TurboV2). On TurboV2, we use MotionClone~\cite{MotionClone} (MC) to align reference motion for the teacher and student models.
To further demonstrate the scalability, \ourname~is also generalized to the DiT distilled model Video-Blade~\cite{Videoblade} (VB), using DiT Wan2.1 with 1.3 billion (Wan2.1-1.3B) parameter variants as the teacher model. All our experiments are conducted using an NVIDIA A100-40G GPU. Experimental configurations and hyperparameters are detailed in the supplementary material.

\paragraph{Datasets.} 
We evaluate our method on two distinct benchmarks. For methods based on TurboV2 and VC2, we remain consistent with prior works~\cite{DMT, MotionInversion}, using source videos obtained from the DAVIS dataset~\cite{Davis}, WebVID dataset~\cite{WebVID}, and online video resources. We refer to the combined validation dataset as TurboBench, consisting of 66 video-edit text pairs derived from 22 unique videos. This benchmark comprehensively spans diverse real-world scenarios, encompassing multiple object categories, varied scene configurations, and rich motion patterns ranging from simple translational movements to complex non-rigid deformations.
For methods based on Wan2.1-1.3B and VB, we extracted 34 real-world videos from the DAVIS dataset~\cite{Davis} for evaluation. Each video is paired with text prompts at three difficulty levels (easy, medium, hard). We refer to this comprehensive evaluation set as DavisBench. We conduct additional evaluation across different benchmarks \cite{LOVEU-TGVE-2023} in the supplementary material.

\paragraph{Metrics.}
To ensure a comprehensive assessment, we perform an evaluation from multiple perspectives.
(1) \textbf{Text Alignment (TA)}: The metric measures how well the generated video content corresponds to the input text prompt by calculating the average cosine similarity between the embeddings of all video frames and the text embeddings, using the CLIP model~\cite{CLIP}.
(2) \textbf{Temporal Consistency (TC)}: The metric evaluates the visual smoothness and coherence of generated videos. We compute cosine similarities between the CLIP image embeddings~\cite{CLIP} across frames and aggregate results into a stability score, which reflects flicker artifacts and semantic discontinuities while allowing motion-driven appearance changes.
(3) \textbf{Motion Fidelity (MF)}: We assess the effectiveness of motion transfer using the Motion Fidelity Score~\cite{DMT}. The metric relies on motion trajectories extracted via Co-Tracker~\cite{CoTracker} and computes the geometric consistency between the motion patterns in the source and generated videos.
(4) \textbf{Fr\'echet Inception Distance (FID)}: We construct a reference image set from selected source videos and evaluate the quality of our generated videos using FID~\cite{FID}.
(5) \textbf{Time Cost}: This measures the total time required to complete motion transfer from a reference video, termed Inference Time Cost (ITC). To ensure a fair comparison, additional fine-tuning or optimization steps are taken into account, termed Training Time Cost (TTC).

\subsection{Comparison with Existing Methods}
To verify the effectiveness of our approach, we compare \ourname~with training-based and training-free methods. The former comprises ControlVideo~\cite{zhao2023controlvideo}, Control-A-Video~\cite{ControlAVideo}, MotionDirector~\cite{MotionDirector}, and MotionInversion~\cite{MotionInversion}. The latter includes MotionClone~\cite{MotionClone}, MOFT~\cite{MOFT}, DMT~\cite{DMT}, and DiTFlow~\cite{DitFlow}. 

\begin{table}[t]
    \caption{Quantitative comparison of VideoCrafter2-based methods on TurboBench.}
    \centering
    \setlength{\tabcolsep}{3mm}
    \resizebox{1.0\linewidth}{!}{
        \begin{tabular}{l c cccccc}
            \toprule
            Method & Backbone & TC $(\uparrow)$ & TA $(\uparrow)$ & MF $(\uparrow)$ &FID $(\downarrow)$ & ITC $(\downarrow)$ &TTC $(\downarrow)$\\
            \midrule
            ControlVideo~\cite{zhao2023controlvideo} &  ControlNet & 0.9420 & 0.251  & 0.964   & 379.37 & 80s & 450s \\
            Control-A-Video~\cite{ControlAVideo} &  ControlNet & 0.925 & 0.256 & 0.858 & 383.95 & 20s & hours \\
            \cmidrule(lr){1-8}
            MotionDirector~\cite{MotionDirector} & VideoCrafter2 & 0.922 & 0.329  & 0.851 & 373.21 & 10s & 280s \\
            MotionInversion~\cite{MotionInversion} & VideoCrafter2 & 0.951  & 0.321  & 0.885  & 351.72  & 32s & 489s \\
            DMT~\cite{DMT} & VideoCrafter2 & 0.961 & 0.259 & 0.909  & 367.18 & 316s & - \\
            MotionClone~\cite{MotionClone} & VideoCrafter2 & \textbf{0.978} & 0.335 & 0.876 & 369.49 & 114s & - \\
            \cmidrule(lr){1-8}
            Ours (MC 16 steps) & TurboV2 & \underline{0.976}  & \textbf{0.348}  & \textbf{0.933}   & \textbf{322.97}  & 13s & -\\
            Ours (MC 8 steps) & TurboV2 & 0.967  & \underline{0.338}  & \underline{0.931}   & \underline{335.65}  & 9s & -\\
            Ours (MC 4 steps) & TurboV2 & 0.956 & 0.323   & 0.927   & 347.91  & 6s & -\\
            \bottomrule
        \end{tabular}
    }
    \label{tab:quantt2vturbov2}  
\end{table}

\begin{table}[t]
    \caption{Quantitative comparison of Wan2.1-1.3B-based methods on DavisBench.}
    \centering
    \setlength{\tabcolsep}{3mm}
    \resizebox{1.0\linewidth}{!}{
        \begin{tabular}{l c cccccc}
            \toprule
            Method & Backbone & TC $(\uparrow)$ & TA $(\uparrow)$ & MF $(\uparrow)$ &FID $(\downarrow)$ & ITC &TTC $(\downarrow)$ \\
            \midrule
            MotionDirector* &Wan2.1-1.3B &0.924 &0.278 &0.627 &135.571 &20s &3629s \\
            MOFT~\cite{MOFT} & Wan2.1-1.3B & 0.961 &\textbf{0.326} & 0.505 & 131.98 & 116s &- \\
            DMT~\cite{DMT} & Wan2.1-1.3B & 0.963 & 0.321 & 0.622  &\textbf{129.49} & 117s &- \\
            DiTFlow~\cite{DitFlow}  & Wan2.1-1.3B & \underline{0.969} &\underline{0.324} & 0.605  & 134.42 & 120s &- \\
            \midrule
            Ours (MOFT 8 steps) & VB & 0.962 & 0.312 & 0.731  &\underline{130.01} & 50s  &-\\
            Ours (DMT 8 steps) & VB & 0.963 & 0.313 &\underline{0.732}  & 132.52 & 49s  &-\\
            Ours (DitFlow 8 steps) & VB &\textbf{0.971} & 0.311 &\textbf{0.733} & 131.82 & 54s &-\\
            \bottomrule
        \end{tabular}
    }
    \label{tab:quantwan}  
\end{table}

\begin{figure}[t]
    \centering
\includegraphics[width=1.0\linewidth]{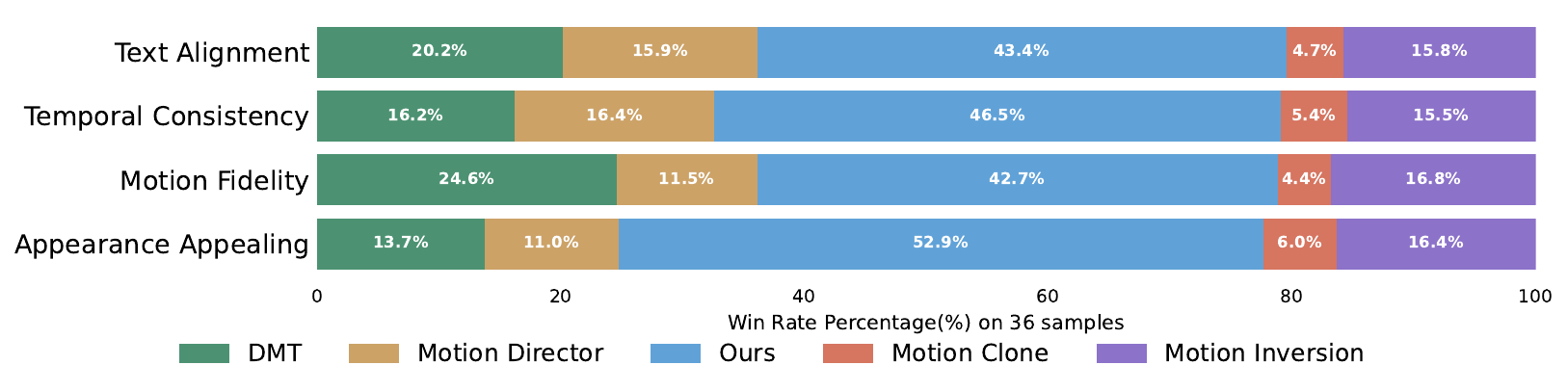}
    \caption{User study results.}
    \label{fig:userstudy}
\end{figure}

\begin{figure*}[t]
    \centering
\includegraphics[width=1.0\linewidth]{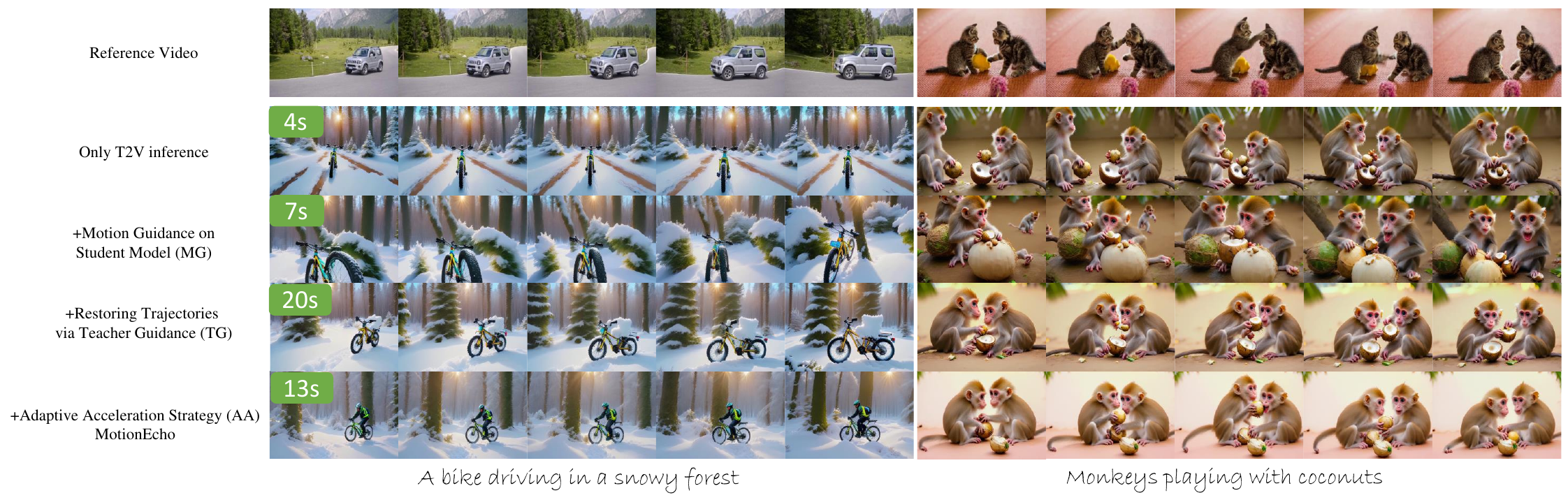}
    \caption{Visualization of the ablation study results.}
    \label{fig:ablation}
\end{figure*}

\paragraph{Quantitative Analysis.}
To objectively evaluate the performance of our method, we conduct quantitative comparisons on two benchmarks using different base models. As shown in Table~\ref{tab:quantt2vturbov2}, our method (MC 16 steps) achieves the best performance across text alignment, motion fidelity, and FID score, while maintaining competitive temporal consistency within just 13 seconds. Compared to prior methods, Control-A-Video~\cite{ControlAVideo} and MotionDirector~\cite{MotionDirector} show similar or higher inference times but significantly lower scores in key quality metrics and require costly training. Moreover, our method can reduce the inference steps while remaining highly competitive. With only 8 steps, it outperforms most baselines across all metrics in just 9 seconds. Even at 4 steps, it achieves a FID of 347.91 and maintains solid temporal consistency, reducing inference time to only 6 seconds. 
Additionally, we apply our method to the Wan2.1-1.3B-based distilled video model (VB), as shown in Table~\ref{tab:quantwan}. Our approach consistently outperforms the training-free baselines (\eg, MOFT, DMT, DiTFlow) on MF, while accelerating the inference process by more than 2×. Furthermore, compared to the training-based baseline MotionDirector* (reproduced on the Wan2.1-1.3B), our method completely bypasses the 3629s training time cost and still delivers substantially better temporal consistency, text alignment, and motion fidelity.

\paragraph{Qualitative Evaluation.}
We separately evaluate our method through qualitative comparisons against baselines based on VideoCrafter2 and Wan2.1-1.3B frameworks. 
In Fig.~\ref{fig:compare1}, compared to the over-smoothing and structural collapse observed in prior motion customization models, our method preserves spatial details while achieving high motion fidelity, particularly under challenging scenarios such as zoom out, snowy forests, and complex multi-object scenes. 
Fig.~\ref{fig:compare2} demonstrates our method achieves consistent improvements on Wan2.1-1.3B-based motion guidance methods. Compared to MOFT, DMT, and DiTFlow, which struggle with motion preservation, our method maintains high fidelity and better visual quality across various motions.
These visual results further validate our method is architecture-agnostic and effective in achieving unified and high-fidelity motion transfer, while maintaining efficiency.

\paragraph{User Study.}
We perform a user study to subjectively assess our method in comparison with VideoCrafter2-based methods, including DMT~\cite{DMT}, MotionClone~\cite{MotionClone}, MotionInversion~\cite{MotionInversion}, and MotionDirector~\cite{MotionDirector}. We randomly sample 36 video-edit text pairs from the TurboBench dataset and synthesize motion videos with the aforementioned methods. 
We invite 50 participants and ask them to compare all generated videos from four different aspects.
Fig.~\ref{fig:userstudy} reports win rates across four perceptual dimensions. MotionEcho achieves the highest preference in all categories, with particularly strong margins in appearance appeal (52.9\%) and temporal consistency (46.5\%), suggesting that teacher-guided trajectory correction produces videos that are not only more motion-accurate but also more visually coherent to human observers.

\begin{table}[t]
    \caption{Ablation study of the key components in our method. \textbf{MG}: Motion Guidance, \textbf{TG}: Teacher Guidance, \textbf{AA}: Adaptive Acceleration.}
    \centering
    \setlength{\tabcolsep}{2.5mm}
    \resizebox{1.0\linewidth}{!}{
    \begin{tabular}{l cccc ccccc}
        \toprule
        Base Model &Backbone & MG & TG & AA & TC $(\uparrow)$ & TA $(\uparrow)$ & MF $(\uparrow)$ & TTC & ITC $(\downarrow)$ \\
        \midrule
        MotionDirector* &TurboV2 & - & - & - & 0.965 & 0.327 & 0.806 &300s & 4s \\
        \midrule
        \multirow{4}{*}{Ours}  
         &TurboV2 & $\times$     & $\times$     & $\times$     & 0.973 & 0.348 & 0.562 & - & 4s \\
         &TurboV2 & $\checkmark$ & $\times$     & $\times$     & 0.967 & 0.324 & 0.833 & - & 7s \\
         &TurboV2 & $\checkmark$ & $\checkmark$ & $\times$     & 0.965 & 0.336 & 0.912 & - & 20s \\
         &TurboV2 & $\checkmark$ & $\checkmark$ & $\checkmark$ & \textbf{0.976} & \textbf{0.348} & \textbf{0.933} & - & 13s \\
        \bottomrule
    \end{tabular}
    }
    \label{tab:ablation} 
\end{table}

\subsection{Ablation Study}
We ablate each component in \ourname~ to achieve comprehensive assessment based on TurboV2 with TurboBench. The results are detailed in Table~\ref {tab:ablation} and Fig.~\ref {fig:ablation}. First, directly applying Motion Guidance (\textbf{MG}) from MotionClone~\cite{MotionClone} injects motion priors and lifts MF from 0.562 to 0.833, but this coarse gradient intervention slightly degrades temporal consistency and text alignment. Then, adding Teacher Guidance (\textbf{TG}) acts as a restorative echo that corrects trajectory deviations, further boosting MF to 0.912 and recovering TA to 0.336, though at the cost of ITC rising to 20s. Finally, the Adaptive Acceleration (\textbf{AA}) strategy selectively activates and truncates the teacher's denoising iterations, cutting ITC back to 13s while reaching the best overall performance. Notably, our full \ourname~pipeline outperforms the training-based baseline MotionDirector (reproduced on TurboV2) in both quality and motion fidelity while eliminating its 300s training cost. Additional ablations are in the \textbf{supplementary material}.

\section{Conclusion}
\label{sec:discussion}
We present \ourname, a training-free framework for motion customization in fast distilled video generators. We identify a fundamental incompatibility between prior guidance methods and distilled models, and address it via teacher-guided test-time distillation, which corrects the student's sampling trajectory through score interpolation. An adaptive acceleration strategy ensures selective and efficient teacher involvement. Experiments across multiple distilled backbones show that \ourname~achieves state-of-the-art motion fidelity and visual quality while retaining the speed of distilled generation.
In future work, we will explore unpaired teacher models with unshared latent spaces, a challenging open problem in diffusion model distillation, and investigate self-assessment mechanisms enabling more efficient and selective speculative guidance during inference.

\section{Acknowledgments}
\label{sec:Acknowledgments}
The contribution of L. Ou was supported by the Baima Lake Laboratory Joint Funds of the Zhejiang Provincial Natural Science Foundation (No.~\seqsplit{LBMHD24F0300023}) and by the National Natural Science Foundation of China (No.~62373329). The contribution of X. Yu was supported by the Zhejiang Provincial Natural Science Foundation of China (No.~LZ25F030003).

\bibliographystyle{splncs04}
\bibliography{main}

\clearpage
\newpage
\beginappendix

\section*{Appendix Overview}

This supplementary material is organized as follows:
\begin{itemize}
    \item \textbf{Section~\ref{app:DME}: More Details of MotionEcho} \\
    Experimental details, pseudo code of MotionEcho, and details of human evaluation.
    
    \item \textbf{Section~\ref{app:MERME}: More Experimental Results of MotionEcho} \\
    More ablation study results, comprehensive experiments on other base models, including AnimateDiff-Lightning, Wan2.2-T2V-5B-Fastvideo, Wan2.2-T2V-5B-Turbo, and TDM, along with more qualitative results, additional experiments across different benchmarks and computational analysis. 
    
    \item \textbf{Section~\ref{app:derivation_tgmd}: Derivation of Teacher-Guided Motion Distillation}\\
    Detailed derivation of the proposed teacher-guided motion distillation.

    \item \textbf{Section~\ref{app:limit}: Limitations and Future Work} \\
    Discussion of current limitations and challenges in \ourname.
    
    \item \textbf{Section~\ref{app:ethical}: Ethical and Social Impacts} \\
    
\end{itemize}

\section{More Details of MotionEcho}
\label{app:DME}

\subsection{Experimental Details}
\label{appsub:EXPD}

\paragraph{Implementation Details.}
Our proposed framework is architecture-agnostic and can be flexibly integrated into various diffusion-based or flow-matching-based video generation pipelines. Below, we detail the specific configurations for two distinct base models.

For TurboV2, we employ the video generation model videocrafter-2~\cite{Videocrafter2} as the teacher text-to-video generation model. The student UNet is initialized from TurboV2 weights and augmented with a motion guidance embedding.
Given a reference video, we encode it into the VAE latent space and perform a single forward pass through the teacher UNet at timestep $t_{\alpha}=399$.
The temporal self-attention probability maps from decoder output blocks 6--8 are cached as the reference motion representation $\mathcal{A} \left( z_{\alpha}^{\mathrm{ref}} \right)$. Similarly to previous works~\cite{FreeTraj, FreeNoise, MotionClone,MotionDirector,DMT}, the reference video is inverted via 200-step DDIM inversion to obtain the initial latent, which is then blended with random noise.
The student model denoise for $N=16/8/4$ steps with motion guidance applied to the initial $\tau=0.5$ of the sampling. At the student denoising step with motion guidance, we perform online distillation according to Algorithm~\ref{alg:adaptive_distillation}. All videos are generated at a resolution of $320\times512$ with 16 frames at 8 fps. 

For VB, we employ the open-sourced video generation model Wan2.1~\cite{wan} as the teacher text-to-video generation model. The teacher keeps the original Wan2.1 transformer weights, while the student starts from the same weights and fuses VB LoRA adapters for faster inference.
A custom UniPC scheduler is used to maintain dual timestep trajectories for teacher and student (typically 200 and 8 denoising steps, respectively), and to support teacher--student $x_0$ mixing during adaptive distillation. 
Given a reference video, we encode it into VAE latents and perform a single forward pass with null text to extract keys/values from the 15th DiT block. The keys/values are seen as the motion prior that are injected into the transformer during guidance. We support three motion objectives: attention-motion-flow matching (DiTFlow), MOFT, and DMT. Guidance is applied on selected student timesteps, with per-step latent optimization using Adam for 5 inner iterations and a linearly decayed learning rate from $2\times 10^{-3}$ to $1\times10^{-3}$. The teacher uses classifier-free guidance (scale 5.0), while the student uses distilled single-branch prediction. Videos are generated at $480\times832$ resolution with 25 frames, and saved at 8 FPS. 

We further support Wan2.2 with the same training/inference recipe: the overall teacher--student design, guidance objectives, KV injection, scheduler, and optimization strategy are unchanged, while the backbone is Wan2.2-TI2V-5B-Turbo / Wan2.2-TI2V-5B-FastVideo and the teacher model is Wan2.2-TI2V-5B. Videos are generated at $480\times832$ resolution with 49 frames.

\paragraph{Hyperparameters.} Most hyperparameters are fixed across all experiments without careful tuning: the guidance schedule ratio $\tau=0.5$, the moving-average window size $W=2$, and the maximum number of teacher distillation events per sample $N_{\max}=3$. The remaining few control the strength and scheduling of teacher guidance; for ease of deployment, we report the settings used on TurboV2 and VB.  
(1) \textit{TurboV2.} Teacher guidance strength $\lambda=0.3$ for all motion types. Motion guidance strength $\eta=0.1$ for camera and hybrid motion, and $\eta=0.2$ for object motion. Noise blend factor $k=0.05$ for camera motion, and $k=0.3$ for object and hybrid motion. The activation and truncation thresholds $(\delta_1,\delta_2)$ are $(500,500)$ for camera motion, $(500,460)$ for hybrid motion, and $(480,380)$ for object motion.
(2) \textit{VB.} Teacher guidance strength $\lambda=0.5$, with thresholds $\delta_1=20$ and $\delta_2=18$. Adaptive distillation is triggered by the moving average of recent student motion losses, activating teacher guidance at most $N_{\max}=3$ times per sample.

\paragraph{Baseline Methods.} We provide a detailed description of the compared baseline used to assess performance in benchmarks.
\begin{itemize}[label=--]
\item ControlVideo~\cite{zhao2023controlvideo} injects HED boundaries as the control signals and finetunes key-frame and temporal attention to enable full cross-frame interaction with a pre-trained diffusion model, ensuring high-quality and consistent text-driven video editing.

\item Control-A-Video \cite{ControlAVideo} incorporates the first-frame condition and uses reward feedback to improve visual quality and motion consistency.

\item MotionDirector \cite{MotionDirector} achieves motion customization by injecting dual-path LoRAs into a pre-trained text-to-video diffusion model, where spatial LoRAs capture appearance and temporal LoRAs model motion, effectively disentangling the two factors.

\item MotionInversion \cite{MotionInversion} learns explicit motion embeddings from a reference video, which are injected into the temporal modules of the text-to-video diffusion model to guide motion generation while removing appearance bias.

\item MotionClone \cite{MotionClone} extracts motion priors from the temporal attention matrix of a reference video and constructs an energy function to guide the sampling process of the pretrained text-to-video model.

\item DMT \cite{DMT} obtains space-time features from intermediate layer activations of the diffusion model and introduces a handcrafted loss to guide motion customization during inference.

\item MOFT \cite{MOFT} uncovers intrinsic motion-aware features in video diffusion models via PCA and defines a Motion Feature (MOFT) by removing content correlations and filtering motion channels.

\item DiTFlow \cite{DitFlow} extracts AMF from cross-frame attention maps of reference videos processed by a pre-trained DiT as motion representation, and guides the denoising process in a training-free manner by optimizing AMF loss, enabling motion transfer from reference videos to generated videos.

\end{itemize}

\subsection{Pseudo Code of MotionEcho}
To enhance reproducibility and ensure a clearer understanding of our proposed test-time distillation framework, we present the detailed pseudo code in Algorithm~\ref{alg:adaptive_distillation}.
Our designed algorithm describes the overall inference procedure for motion-customized video generation using a distilled student model guided by a fine-grained teacher model.
The algorithm also incorporates an adaptive acceleration strategy, which dynamically determines when and how intensively teacher supervision should be applied, ensuring an optimal trade-off between efficiency and motion fidelity.

\begin{algorithm*}[t]
\caption{MotionEcho on TurboV2}
\label{alg:adaptive_distillation}
\begin{algorithmic}[1]
    \State \textbf{Input:} 
        Distilled video diffusion model $\epsilon_{\psi}$, 
        Fine-grained teacher model $\epsilon_{\theta}$,  
        Reference latent $z^{\text{ref}}$,
        Text prompt $c$, Unconditional prompt $\phi$, 
        Teacher guidance schedule ratio $\tau$, 
        Blend scaling factor $k$,
        Motion guidance strength $\eta$, 
        Teacher guidance strength $\lambda$, 
        Motion thresholds $\delta_1$, $\delta_2$, 
        Window size $W$, Teacher's max inner denoising steps $N_{\text{max}}$, 
        Total student sampling steps $N_s$, Total teacher sampling steps $N_t$.
    \State \textbf{Output:} Final denoised latent $z_0^{\psi}$. \\

    \State $z_T^{\text{ref}}, z_{t_\alpha}^{\text{ref}}$ = DDIM Inversion($z^{\text{ref}}$)
    \State $\epsilon_T \sim \mathcal{N}(0, \mathbf{I})$
    \State Initialize blended noise: $z_T^{\psi} = \sqrt{k} z_T^{\text{ref}} + \sqrt{1-k} \epsilon_T$
    \State Set $\Delta t_s = T / N_s$, $\Delta t_t = T / N_t$
    
    \For{$t = T$ \textbf{to} $\Delta t_s$ \textbf{by} $-\Delta t_s$}
        \State \textbf{Stage 1.} \textit{Student Customization}
        \State $\tilde{\epsilon}_{\psi} (z^{\psi}_{t}, c, t) = (1 + \omega_{\psi}) \epsilon_{\psi}(z^{\psi}_{t}, c, t) - \omega_{\psi} \epsilon_{\psi}(z^{\psi}_{t}, \phi, t)$
        \State $\tilde{z}_{0\leftarrow t}^{\psi} = \frac{z^{\psi}_{t} - \sqrt{1 - \bar{\alpha}_{t}} \tilde{\epsilon}_{\psi} (z^{\psi}_{t}, c, t)}{\sqrt{\bar{\alpha}_{t}}}$ \Comment{Without motion guidance}
        
        \State $\hat{\epsilon}_{\psi} (z^{\psi}_{t}, c, t) = (1 + \omega_{\psi}) \epsilon_{\psi}(z^{\psi}_{t}, c, t) - \omega_{\psi} \epsilon_{\psi}(z^{\psi}_{t}, \phi, t) - \eta \cdot \nabla_{z_t^{\psi}} \mathcal{G}^m(z_t^{\psi}, z_{t_\alpha}^{\text{ref}}, t)$ 
        \State $\hat{z}_{0\leftarrow t}^{\psi} = \frac{z^{\psi}_{t} - \sqrt{1 - \bar{\alpha}_{t}} \hat{\epsilon}_{\psi} (z^{\psi}_{t}, c, t)}{\sqrt{\bar{\alpha}_{t}}}$ \Comment{With motion guidance}
        
        \State \textbf{Stage 2.} \textit{Adaptive Teacher Guidance}
        \If{$t > \tau \cdot T$}
            \State $\text{avg}_{\mathcal{G}}^{\psi}(t) = \frac{1}{W} \sum_{j=0}^{W-1} \mathcal{G}^m(z_{t+j \cdot \Delta t_s }^{\psi}, z_{t_\alpha}^{\text{ref}}, t+j\cdot \Delta t_s)$ \Comment{Moving average motion loss}
            \If{$\text{avg}_{\mathcal{G}}^{\psi}(t) > \delta_1$}
                \State Select inner teacher step $s \in (t, t - \Delta t_s)$
                
                \State $z_s^{\theta} = \sqrt{\bar{\alpha}_s} \hat{z}^\psi_{0 \leftarrow t} + \sqrt{1 - \bar{\alpha}_s} \epsilon, \; \epsilon \sim \mathcal{N}(0, I)$  \Comment{Renoise}
                \For{$n = s$ \textbf{to} $t-\Delta t_s$ \textbf{by} $-\Delta t_t$}
                    \State $\hat{\epsilon}_{\theta}(z_n^{\theta}, c, n) = (1 + \omega_{\theta}) \epsilon_{\theta}(z_n^{\theta}, c, n) - \omega_{\theta} \epsilon_{\theta}(z_n^{\theta}, \phi, n) - \eta \cdot \nabla_{z_n^{\theta}} \mathcal{G}^m(z_n^{\theta}, z_{t_\alpha}^{\text{ref}}, n)$
                    \State $\hat{z}_{0\leftarrow n}^{\theta} = \frac{z^{\theta}_{n} - \sqrt{1 - \bar{\alpha}_{n}} \hat{\epsilon}_{\theta} (z^{\theta}_{n}, c, n)}{\sqrt{\bar{\alpha}_{n}}}$ \Comment{Teacher motion customization}

                    \If{$\mathcal{G}^m(z_n^{\theta}, z_{t_\alpha}^{\text{ref}}, n) < \delta_2$ or {$\frac{s-n}{\Delta t_t}>N_\text{max}$} } 
                        \State $z_{t-\Delta t_s}^{\theta} = \sqrt{\bar{\alpha}_{t-\Delta t_s}} \hat{z}^\theta_{0 \leftarrow n} + \sqrt{1 - \bar{\alpha}_{t-\Delta t_s}} \epsilon, \; \epsilon \sim \mathcal{N}(0, I)$
                    \Else
                        \State $z_{n-\Delta t_t}^{\theta} = \sqrt{\bar{\alpha}_{n-\Delta t_t}} \hat{z}^\theta_{0 \leftarrow n} + \sqrt{1 - \bar{\alpha}_{n-\Delta t_t}} \epsilon, \; \epsilon \sim \mathcal{N}(0, I)$
                    \EndIf
                \EndFor

                \State $\hat{z}_{0\leftarrow {t-\Delta t_s}}^{\theta} = \frac{z^{\theta}_{t-\Delta t_s} - \sqrt{1 - \bar{\alpha}_{t-\Delta t_s}} \hat{\epsilon}_{\theta} (z^{\theta}_{t-\Delta t_s}, c, t-\Delta t_s)}{\sqrt{\bar{\alpha}_{t-\Delta t_s}}}$  \Comment{One-step prediction}

                \State $\hat{z}_{0 \leftarrow t}^{\text{new}} = (1 - \lambda) \hat{z}_{0\leftarrow t}^{\psi} + \lambda \hat{z}_{0\leftarrow {t-\Delta t_s}}^{\theta}$
            \Else
                \State $\hat{z}_{0 \leftarrow t}^{\text{new}} = \hat{z}_{0 \leftarrow t}^{\psi}$
            \EndIf
        \Else
            \State $\hat{z}_{0 \leftarrow t}^{\text{new}} = \tilde{z}_{0 \leftarrow t}^{\psi}$
        \EndIf

        \State \textbf{Stage 3.} \textit{Update latent}
        \State $z_{t-\Delta t_s}^{\psi} = \sqrt{\bar{\alpha}_{t - \Delta t_s}} \cdot \hat{z}_{0 \leftarrow t}^{\text{new}} + \sqrt{1 - \bar{\alpha}_{t - \Delta t_s}} \cdot \epsilon, \; \epsilon \sim \mathcal{N}(0, I)$
    \EndFor
    \State \Return $z_0^{\psi}$
\end{algorithmic}
\end{algorithm*}

\subsection{Details on Human Evaluation}
\begin{figure*}[t]
    \centering
    \includegraphics[height=0.75\textheight, keepaspectratio]{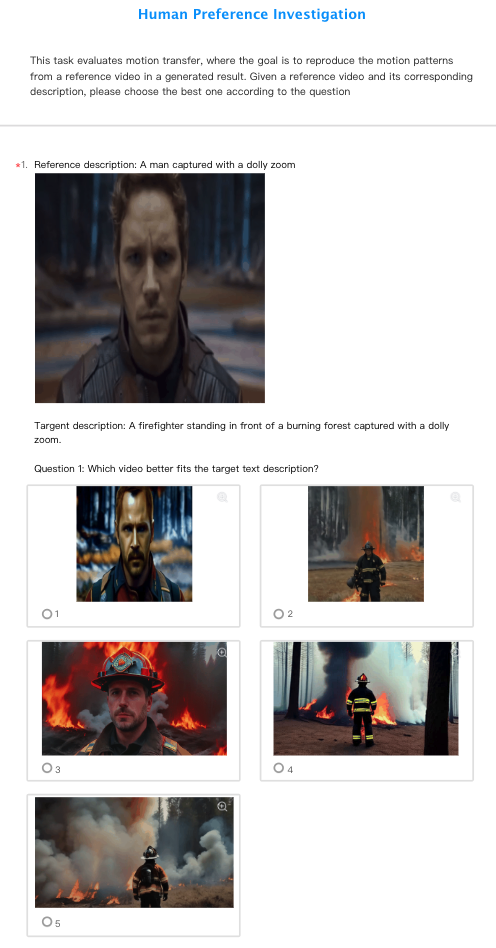}
    \caption{The screenshot of human preference investigation: Which video better fits the target text description?}
    \label{fig:Q1}
\end{figure*}

\begin{figure*}[t]
    \centering
    \includegraphics[height=0.75\textheight, keepaspectratio]{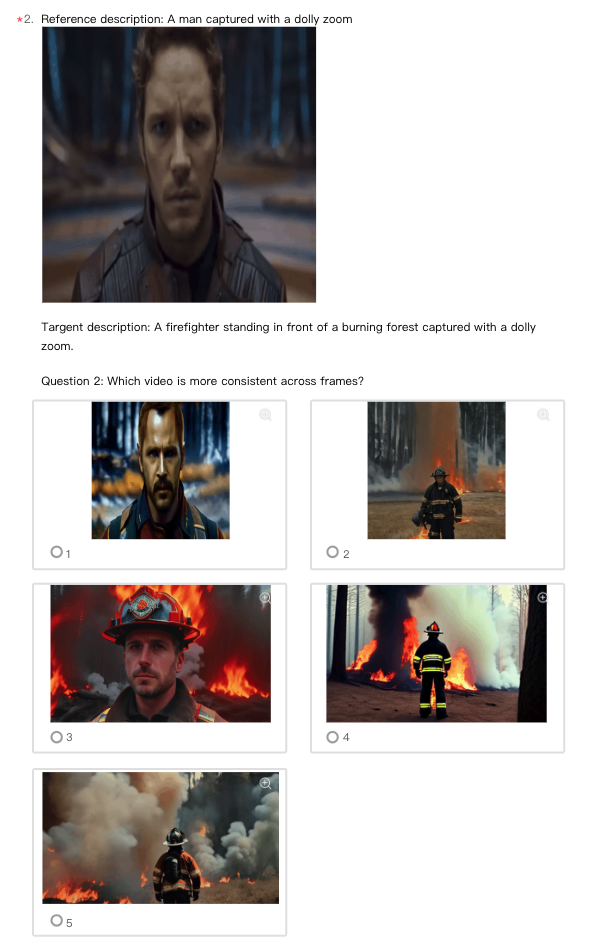}
    \caption{The screenshot of human preference investigation: Which video is more consistent across frames?}
    \label{fig:Q2}
\end{figure*}

\begin{figure*}[t]
    \centering
    \includegraphics[height=0.75\textheight, keepaspectratio]{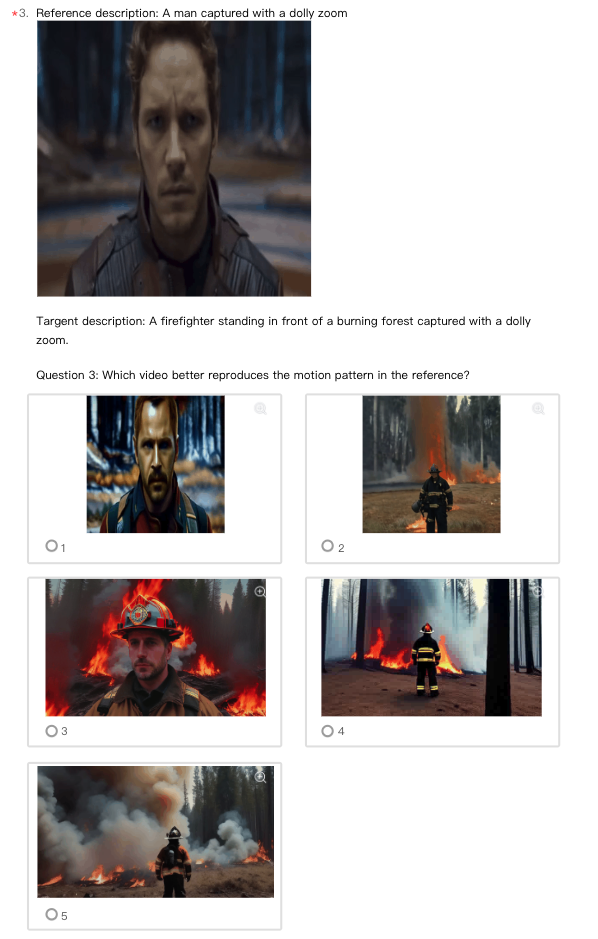}
    \caption{The screenshot of human preference investigation: Which video better reproduces the motion pattern in the reference?}
    \label{fig:Q3}
\end{figure*}

\begin{figure*}[t]
    \centering
    \includegraphics[height=0.75\textheight, keepaspectratio]{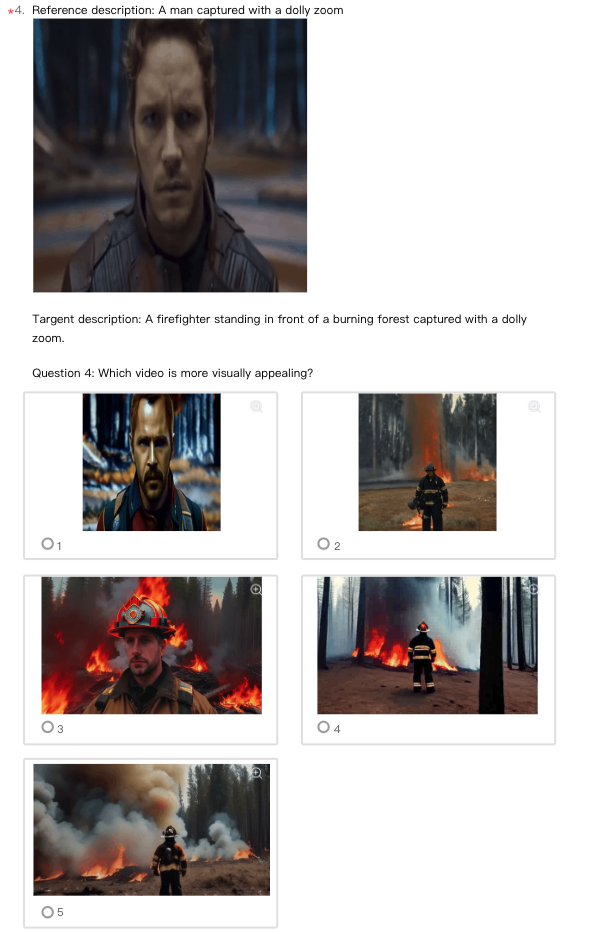}
    \caption{The screenshot of human preference investigation: Which video is more visually appealing?}
    \label{fig:Q4}
\end{figure*}

We conduct a user study to assess the perceptual quality of motion-customized videos generated by \ourname. Specifically, the study is implemented via an online survey form, which is divided into multiple sections. Each section corresponds to a specific test case consisting of a reference video, a given prompt and generated videos by our method and baselines. Within each section, investigators are required to answer four questions (Q1–Q4), corresponding to Figure~\ref{fig:Q1}, Figure~\ref{fig:Q2}, Figure~\ref{fig:Q3} and Figure~\ref{fig:Q4}. In addition, investigators are recruited through an online platform, ensuring their anonymity. Each investigator is required to have at least a bachelor’s degree and their privacy and identity are kept confidential throughout the entire process. Finally, the collected responses are aggregated to perform comparative analysis across different methods in terms of user-perceived quality.

\section{More Experimental Results of MotionEcho}
\label{app:MERME}

\subsection{Ablation Study Results}

Here, we present additional ablation study results on TurboV2. 

\paragraph{Effect of Teacher Guidance Strength $\lambda$.} Table~\ref{tab:ablation_guidance}~(left) shows how varying the teacher guidance strength $\lambda$ influences the generated video quality. We observe that $\lambda = 0.3$ offers the best overall trade-off across three metrics.

\paragraph{Effect of Motion Guidance Strength $\eta$.} As shown in Table~\ref{tab:ablation_guidance}~(right), motion guidance strength $\eta$ controls how strictly the model follows the reference motion. Increasing $\eta$ improves motion fidelity but reduces textual alignment or introduces temporal artifacts when overly strong.

\paragraph{Effect of guidance activation threshold $\delta_1$.}
The teacher guidance activation threshold $\delta_1$ controls when the teacher model is invoked during the distillation process. 
We evaluate $\delta_1$ across a range from 360 to 500 with an interval of 20. As shown in Table~\ref{tab:ablation_thresholds}~(left), increasing $\delta_1$ from 360 to 500 significantly reduces inference time from 19.89s to 13.29s, while motion fidelity remains stable with minimal impact on text alignment and temporal consistency.
This demonstrates that our method is robust to the choice of $\delta_1$, allowing users to flexibly adjust this threshold to balance inference speed and quality.

\paragraph{Effect of motion guidance truncation threshold $\delta_2$ for teacher model.} 
{Once the teacher's motion loss falls below a confidence threshold $\delta_2$, we end the motion transfer early and continue denoising for efficiency.} 
As shown in Table~\ref{tab:ablation_thresholds}~(right), increasing $\delta_2$ from 380 to 500 reduces inference time from 17.83s to 12.84s.
Importantly, our acceleration comes with negligible impact on motion fidelity and even slight improvements in text alignment.
These results confirm that our method is not overly sensitive to $\delta_2$, achieving a favorable balance between efficiency and quality across different hyperparameter settings.

\begin{table}[t]
    \caption{Ablation studies on the teacher guidance strength $\lambda$ (left) and motion guidance strength $\eta$ (right).}
    \label{tab:ablation_guidance}
    \centering
    \begin{minipage}{0.48\textwidth}
        \centering
        \setlength{\tabcolsep}{2mm}
        \resizebox{\linewidth}{!}{
        \begin{tabular}{ccccc}
            \toprule
            $\lambda$ & TC $(\uparrow)$ & TA $(\uparrow)$ & MF $(\uparrow)$ & FID $(\downarrow)$ \\
            \midrule
            0.1 & 0.975 & 0.341 & 0.924 & 332.23 \\
            0.2 & 0.975 &0.340 &0.929 & 327.18 \\
            0.3 & \textbf{0.977} & 0.344 & \textbf{0.936} & 323.91 \\
            0.4 &0.976 &0.351 &0.930 &321.84 \\
            0.5 & 0.976 & \textbf{0.359} & 0.930 & \textbf{320.10} \\
            \bottomrule
        \end{tabular}
        }
    \end{minipage}\hfill
    \begin{minipage}{0.48\textwidth}
        \centering
        \setlength{\tabcolsep}{2mm}
        \resizebox{\linewidth}{!}{
        \begin{tabular}{ccccc}
            \toprule
            $\eta$ & TC $(\uparrow)$ & TA $(\uparrow)$ & MF $(\uparrow)$ & FID $(\downarrow)$ \\
            \midrule
            0.05 & \textbf{0.978} & \textbf{0.350} & 0.929 & \textbf{323.91} \\
            0.1  & 0.973 & 0.341 & 0.935 & 327.77 \\
            0.15 &0.970 &0.339 &0.936 &326.21 \\
            0.2  & 0.966 & 0.338 & \textbf{0.940} & 325.17 \\
            \bottomrule
        \end{tabular}
        }
    \end{minipage}
\end{table}

\begin{table}[t]
    \caption{Ablation studies on the teacher guidance activation threshold $\delta_1$ (left) and motion guidance truncation threshold $\delta_2$ (right).}
    \label{tab:ablation_thresholds}
    \centering
    \begin{minipage}{0.48\textwidth}
        \centering
        \setlength{\tabcolsep}{1.5mm}
        \resizebox{\linewidth}{!}{
        \begin{tabular}{ccccc}
            \toprule
            $\delta_1$ & MF $(\uparrow)$ & TA $(\uparrow)$ & TC $(\uparrow)$ & ITC $(\downarrow)$ \\
            \midrule
            360 & \textbf{0.9314} & 0.3392 & 0.9724 & 19.89 \\
            380 & 0.9300 & 0.3373 & \textbf{0.9730} & 17.20 \\
            400 & 0.9306 & \textbf{0.3396} & 0.9726 & 17.12 \\
            420 & 0.9298 & 0.3376 & 0.9723 & 16.11 \\
            440 & 0.9285 & 0.3373 & 0.9714 & 15.12 \\
            460 & 0.9297 & 0.3387 & 0.9721 & 15.60 \\
            480 & 0.9284 & 0.3384 & 0.9721 & 13.92 \\
            500 & 0.9308 & 0.3380 & 0.9724 & \textbf{13.29} \\
            \bottomrule
        \end{tabular}
        }
    \end{minipage}\hfill
    \begin{minipage}{0.48\textwidth}
        \centering
        \setlength{\tabcolsep}{1.5mm}
        \resizebox{\linewidth}{!}{
        \begin{tabular}{ccccc}
            \toprule
            $\delta_2$ & MF $(\uparrow)$ & TA $(\uparrow)$ & TC $(\uparrow)$ & ITC $(\downarrow)$ \\
            \midrule
            360 & 0.9319 & 0.3395 & 0.9729 & 17.56 \\
            380 & \textbf{0.9326} & 0.3396 & 0.9724 & 17.83 \\
            400 & 0.9317 & 0.3401 & 0.9724 & 16.73 \\
            420 & 0.9297 & 0.3386 & 0.9726 & 15.48 \\
            440 & 0.9315 & 0.3382 & \textbf{0.9731} & 14.44 \\
            460 & 0.9314 & \textbf{0.3408} & 0.9729 & 13.78 \\
            480 & 0.9307 & 0.3402 & 0.9728 & 13.06 \\
            500 & 0.9303 & 0.3400 & 0.9723 & \textbf{12.84} \\
            \bottomrule
        \end{tabular}
        }
    \end{minipage}
\end{table}

\subsection{Experiments on Additional Base Models}
\begin{figure*}[t]
    \centering
    \includegraphics[width=1.0\linewidth]{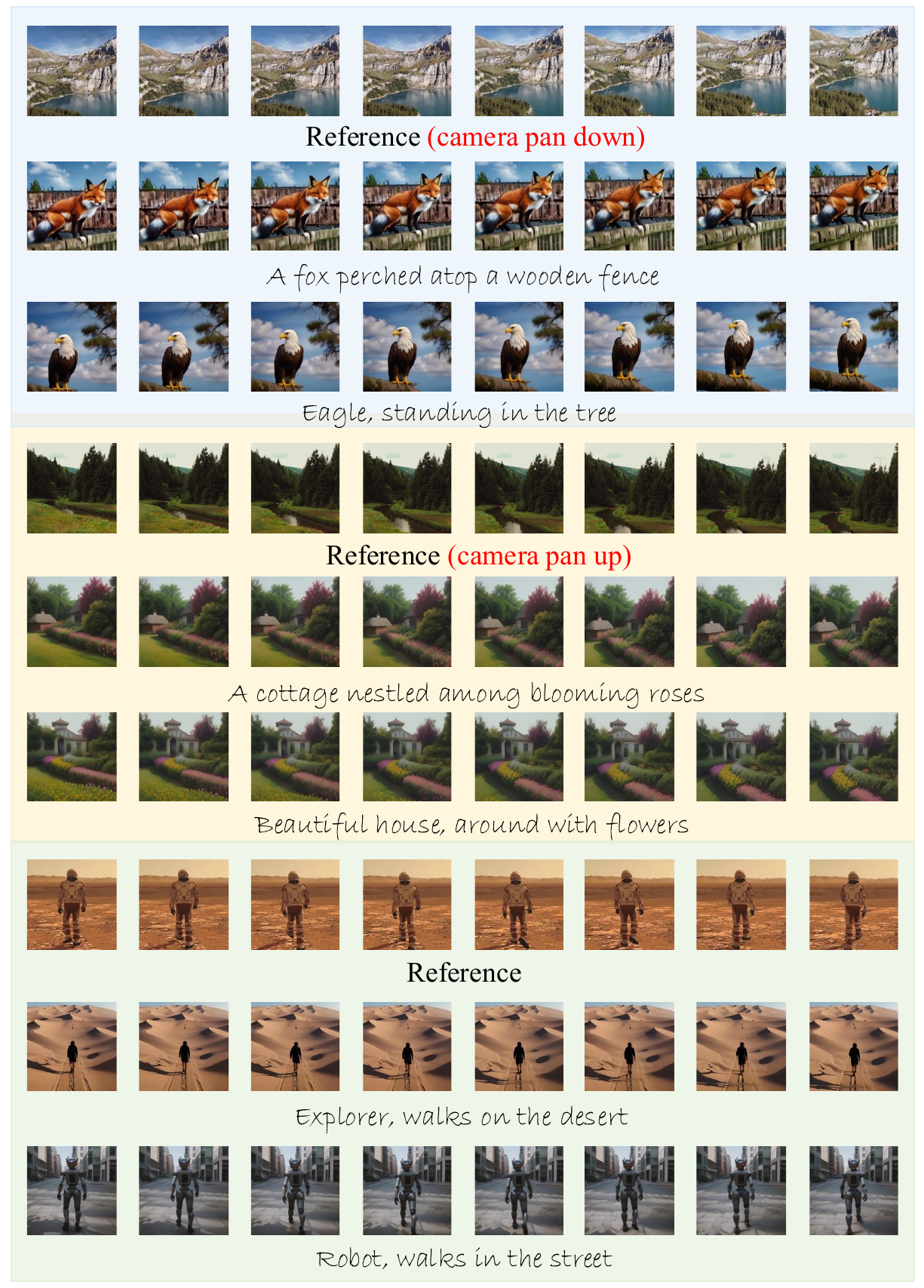}
    \caption{More results of our method (MotionClone+AnimateDiff-Lightning) in camera and object motion customization.}
    \label{fig:dADL}
\end{figure*}

\paragraph{Animatediff-Lightning.}
To further demonstrate the generalization and scalability of our method, \ourname ~is generalized to the distilled model AnimateDiff-Lightning \cite{AnimateDiffLightning} (AD-L), using AnimateDiff \cite{AnimateDiff} (AD) as the teacher model. 
AD-L significantly accelerates the generation process through progressive distillation, reducing the required sampling steps while maintaining comparable generation quality.
Additionally, we use MotionClone \cite{MotionClone} (MC) and MotionDirector (MD) \cite{MotionDirector} to align reference motion for the teacher and student models.
For methods based on AD and AD-L, we make a thorough analysis on 15 real videos from MotionClone~\cite{MotionClone}, comprising 7 videos with camera motion and 8 videos for object motion, hereafter referred to as AnimateBench. 
The results in Table~\ref{tab:quantanimatediff} further verify the effectiveness, superiority and flexibility of our method.
In Figure~\ref{fig:dADL}, our method based on AD-L shows better motion alignment and temporal consistency than both MotionClone \cite{MotionClone} and MotionDirector \cite{MotionDirector}. For example, our method handles subtle motions (\eg, head turning) and large camera movements (\eg, rotating island) more effectively, with fewer artifacts and less drift.

\paragraph{Wan2.2-TI2V-5B-Fastvideo and Wan2.2-TI2V-5B-Turbo.}
To validate the effectiveness of \ourname~in the cutting-edge large-scale video generation model, we extend \ourname~to the distilled model, Wan2.2-TI2V-5B-Fastvideo, using Wan2.2-TI2V-5B as the teacher model.
As presented in Table~\ref{tab:quantwan22}, Figure~\ref{fig:c12} and Figure~\ref{fig:c13}, deploying existing training-free methods (MOFT, DMT, DiTFlow) directly on the full teacher model achieves limited motion fidelity and suffers from severe computational bottlenecks (ITC $>$ 120 s). By integrating \ourname~with the 3-step Fastvideo, we achieve a substantial 2.5× inference acceleration (ITC $\approx$ 50s) while achieving significant improvements in motion transfer. Notably, \ourname~(DiTFlow 3 steps) on Fastvideo pushes the MF to 0.814, comprehensively outperforming the teacher baseline. Furthermore, parallel evaluations on the Wan2.2-TI2V-5B-Turbo exhibit equally robust motion control capabilities, collectively proving that \ourname~remains highly effective on different distilled model variants.

\paragraph{TDM.}
\ourname ~is also generalized to the distilled DiT model TDM\cite{TDM} which has different architecture and scales, using CogVideoX-2B~\cite{cogvideox} as the teacher model. 
We use MOFT~\cite{MOFT}, DMT~\cite{DMT} and DiTFlow~\cite{DitFlow} to transfer reference motion, and evaluate them on DavisBench. As shown in Table~\ref{tab:quantcogvideox}, applying \ourname~to CogVideoX-2B-TDM achieves strong performance across all three guidance variants while substantially reducing inference time. 
Compared with the teacher-based counterparts, our method reduces ITC from 82--90s to 47--50s, while maintaining comparable or even better generation quality. In particular, \ourname~achieves the best TC of 0.970 with DiTFlow and the best TA of 0.343 with DMT, while also reaching the highest MF of 0.726 with DiTFlow. 

\paragraph{Video-Blade.} 
{For a complete \textbf{baseline comparison on VB settings}, we directly apply MOFT, DMT and DiTFlow on VB without any teacher guidance. The results in Table.~\ref{tab:quantwanvb} show that our method consistently recovers motion fidelity with only moderate overhead, verifying the effectiveness of our method.}

\begin{table}[t]
    \caption{Quantitative comparison of AnimateDiff-based methods on AnimateBench.}
    \centering
    \setlength{\tabcolsep}{3mm}
    \resizebox{0.98\linewidth}{!}{
        \begin{tabular}{cccccccc}
            \toprule
            Method &Backbone & TC $(\uparrow)$ & TA $(\uparrow)$ & MF $(\uparrow)$ &FID $(\downarrow)$ & ITC $(\downarrow)$ &TTC $(\downarrow)$\\
            \midrule
            MotionDirector \cite{MotionDirector} &AnimateDiff &0.961 &0.304  &0.833 &378.31 &18s &451s \\
            MotionClone \cite{MotionClone} &AnimateDiff &0.957 &0.321 &0.849  &367.07 &203s &- \\
            MC+AD-L (8steps) \cite{MotionClone} &AnimateDiff-Lightning &0.974 &\underline{0.322} &0.648  &357.15 &7s &- \\
            \midrule
            Ours (MC+AD-L 8 steps) &AnimateDiff-Lightning &\textbf{0.981} &\textbf{0.327} &\textbf{0.868} &\textbf{336.03} &24s & -\\
            Ours (MC+AD-L 4 steps) &AnimateDiff-Lightning &\underline{0.973} &0.319 &\underline{0.854} &\underline{348.99} &17s &-\\
            \bottomrule
        \end{tabular}
    }
    \label{tab:quantanimatediff}  
\end{table}

\begin{table}[t]
    \caption{Quantitative comparison of CogVideoX-2b on DavisBench.}
    \centering
    \setlength{\tabcolsep}{3mm}
    \resizebox{1.0\linewidth}{!}{
        \begin{tabular}{ccccccc}
            \toprule
            Method &Backbone & TC $(\uparrow)$ & TA $(\uparrow)$ & MF $(\uparrow)$ &FID $(\downarrow)$ & ITC $(\downarrow)$ \\
            \midrule
            MOFT \cite{MOFT} &CogVideoX-2b &0.958 &0.311  &0.485 &134.88 &84s  \\
            DMT \cite{DMT} &CogVideoX-2b &0.954 &0.332 &{0.715}  &126.14 &82s  \\
            DiTFlow \cite{DitFlow}  &CogVideoX-2b &0.962 &{0.338} &0.720  &130.54 &90s  \\
            \midrule
            Ours (MOFT 4 steps) &CogVideoX-2b-TDM &0.969 &0.342 &0.528  &127.52 &48s \\
            Ours (DMT 4 steps) &CogVideoX-2b-TDM &0.968 &\textbf{0.343} &0.700  &126.24 &47s  \\
            Ours (DiTFlow 4 steps) &CogVideoX-2b-TDM &\textbf{0.970} &0.340 &\textbf{0.726} &128.11  &50s \\
            \bottomrule
        \end{tabular}
    }
    \label{tab:quantcogvideox}  
\end{table}

\begin{table}[t]
    \caption{Quantitative comparison of Wan2.2-TI2V-5B on DavisBench.}
    \centering
    \setlength{\tabcolsep}{3mm}
    \resizebox{1.0\linewidth}{!}{
        \begin{tabular}{ccccccc}
            \toprule
            Method &Backbone & TC $(\uparrow)$ & TA $(\uparrow)$ & MF $(\uparrow)$ &FID $(\downarrow)$ & ITC $(\downarrow)$ \\
            \midrule
            MOFT \cite{MOFT} &Wan2.2TI2V-5B &0.971 &0.316  &0.590 &139.32 &124s  \\
            DMT \cite{DMT} &Wan2.2TI2V-5B &0.972 &0.318  &0.579  &138.37 &122s  \\
            DiTFlow \cite{DitFlow}  &Wan2.2TI2V-5B &0.974 &0.316  &0.570 &136.19  &130s  \\
            \midrule
            Ours (MOFT 3 steps) &Wan2.2TI2V-5B-Fastvideo &0.978 &0.324 &0.799 &138.49&48s \\
            Ours (DMT 3 steps)  &Wan2.2TI2V-5B-Fastvideo &0.977 &0.324 &0.804  &137.42 &47s  \\
            Ours (DiTFlow 3 steps) &Wan2.2TI2V-5B-Fastvideo &0.975 &0.321 &0.814 &138.90  &50s \\
            \midrule
            Ours (MOFT 4 steps) &Wan2.2TI2V-5B-Turbo &0.955 &0.293 &0.751  &142.62 &51s \\
            Ours (DMT 4 steps)  &Wan2.2TI2V-5B-Turbo &0.955 &0.287 &0.777  &142.29 &48s  \\
            Ours (DitFlow 4 steps) &Wan2.2TI2V-5B-Turbo &0.957 &0.290 &0.772 &145.29  &55s \\

            \bottomrule
        \end{tabular}
    }
    \label{tab:quantwan22}  
\end{table}

\begin{table}[t]
\centering
\caption{
Per-type MF and Flow-sim (cos similarity between RAFT-estimated flows of reference and generated videos, robust to global camera transformations) on AnimateBench.}
\centering
\label{tab:pertype}
\setlength{\tabcolsep}{5mm}
\resizebox{\linewidth}{!}{
\begin{tabular}{c|ccc|cc|c}
\hline
Method & Cam. MF$\uparrow$ & Obj. MF$\uparrow$ & MF$\uparrow$ & Cam. Flow-sim$\uparrow$ &Obj. Flow-sim$\uparrow$ & ITC $\downarrow$ \\
\hline
MotionClone &\textbf{0.940} &0.759 &0.849 &0.841 &0.548 &203s\\
MC+AD-L (8 steps) & 0.717 & 0.579 & 0.648 &0.240 &0.136 &7s\\
\textbf{Ours (MC+AD-L 8steps)} & 0.938 & \textbf{0.798} & \textbf{0.868} &\textbf{0.866} &\textbf{0.588} &24s \\
\hline
\end{tabular}
}
\end{table}

\begin{table}[t]
    \caption{Quantitative comparison of VB-only on DavisBench.}
    \centering
    \setlength{\tabcolsep}{5mm}
    \resizebox{1.0\linewidth}{!}{
        \begin{tabular}{c c ccccc}
            \hline
            Method & Backbone & TC $(\uparrow)$ & TA $(\uparrow)$ & MF $(\uparrow)$ &FID $(\downarrow)$ & ITC  \\
            \hline
            MOFT (VB-8steps) & VB & 0.955 & 0.303 & 0.491  & 135.86 & 33s  \\
            DMT (VB-8steps) & VB & 0.954 & 0.302 & 0.567  & 133.28 & 32s   \\
            DitFlow (VB-8steps) & VB & 0.964 & 0.304 & 0.540  & 138.90 & 34s \\
            \hline
            Ours (MOFT 8 steps) & VB & 0.962 & 0.312 & 0.731  &\textbf{130.01} & 50s  \\
            Ours (DMT 8 steps) & VB & 0.963 & \textbf{0.313} &0.732  & 132.52 & 49s \\
            Ours (DitFlow 8 steps) & VB &\textbf{0.971} & 0.311 &\textbf{0.733} & 131.82 & 54s \\
            \hline
        \end{tabular}
    }
    \label{tab:quantwanvb}  
\end{table}

\begin{table}
    \caption{Performance Comparison on LOVEU-TGVE Benchmark.}
    \centering
    \setlength{\tabcolsep}{5mm}
    \resizebox{1.0\linewidth}{!}{
    \begin{tabular}{lccc}
        \toprule
        Method & AD $(\uparrow)$ & TC $(\uparrow)$ & Pick Score $(\uparrow)$ \\
        \midrule
        Tune-a-Video~\cite{TuneAVideo}      & 25.95 & 0.924 & 20.09 \\
        MotionDirector~\cite{MotionDirector}    & 25.95 & 0.931 & 20.37 \\
        PointMotion~\cite{PointMotion} & 25.74 & \textbf{0.933} & 20.34 \\
        \textbf{Ours (MC+TurboV2 16 steps)} & \textbf{28.16} & 0.918 & \textbf{21.41} \\
        \bottomrule
    \end{tabular}
    }
    \label{tab:loveu_tgve}
\end{table}

\subsection{Experiments with Additional Metric}
{Table~\ref{tab:quantanimatediff} reports the motion-type--stratified MF on AnimateBench. To further examine performance across motion categories, we provide per-type MF results in Table~\ref{tab:pertype}, where our method achieves competitive performance on both camera and object motion. For a more comprehensive evaluation, we also include a dense optical-flow metric in Table~\ref{tab:pertype}. Specifically, Flow-sim measures the cosine similarity between RAFT-estimated optical-flow fields of the reference and generated videos, and is robust to global camera transformations. The results show that \ourname~outperforms the baselines on both the camera- and object-motion subsets, indicating that the aggregate MF score does not obscure camera-motion weaknesses.}

\subsection{Experiments with Additional Benchmark}

To further validate the generalizability of our approach, we conduct extensive experiments on the LOVEU-TGVE benchmark~\cite{LOVEU-TGVE-2023} from CVPR 2023's text-guided video editing track. This benchmark comprises 76 videos, each paired with 4 edit prompts, resulting in 304 video-text editing pairs that cover a wide range of appearance and motion editing scenarios. Compared to the 66-pair evaluation, this benchmark offers significantly more diversity and complexity, providing a more comprehensive assessment of robustness in real-world scenarios.

As shown in Table~\ref{tab:loveu_tgve}, our method, \ourname, achieves superior performance across multiple metrics. Specifically, we obtain the highest Appearance Diversity (AD) score of 28.16, substantially outperforming all baseline methods including Tune-a-Video~\cite{TuneAVideo}, MotionDirector~\cite{MotionDirector}, and PointMotion~\cite{PointMotion}. This indicates that our method generates more visually diverse and semantically aligned edits. Moreover, we achieve the best Pick Score of 21.41, demonstrating superior overall quality and user preference compared to competing approaches. Although PointMotion~\cite{PointMotion} achieves slightly higher Temporal Consistency (TC), our method maintains competitive temporal coherence while significantly excelling in AD and overall quality. These results on a larger and more varied benchmark further demonstrate the robustness, effectiveness, and generalizability of \ourname ~in handling complex and diverse real-world video editing tasks.

\subsection{More Qualitative Results}
We provide more visual results for qualitative evaluation.

\begin{figure*}[t]
    \centering    \includegraphics[width=1.0\linewidth]{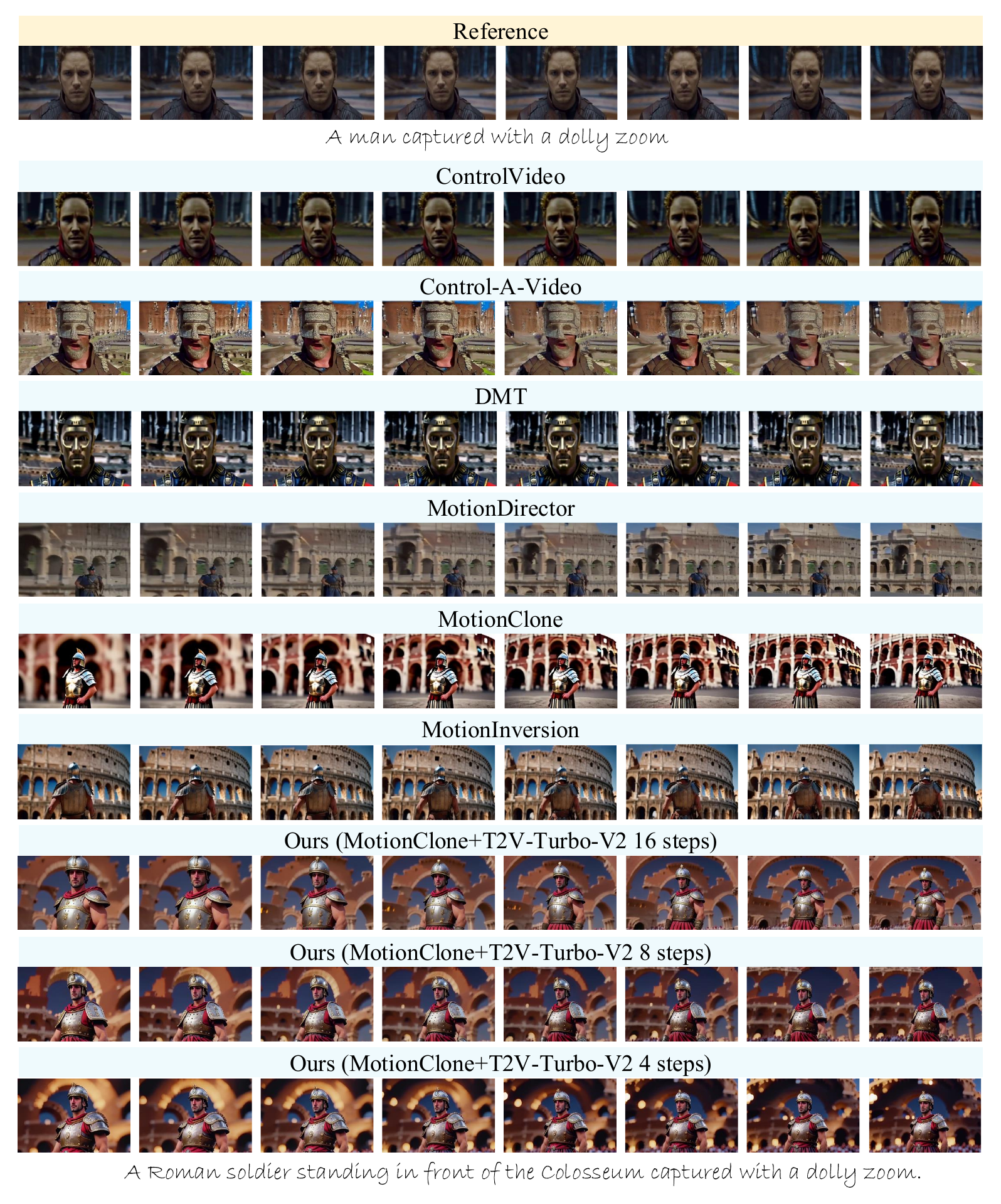}
    \caption{Additional comparison results in camera motion customization.}
    \label{fig:c1}
\end{figure*}

\begin{figure*}[t]
    \centering
    \includegraphics[width=1.0\linewidth]{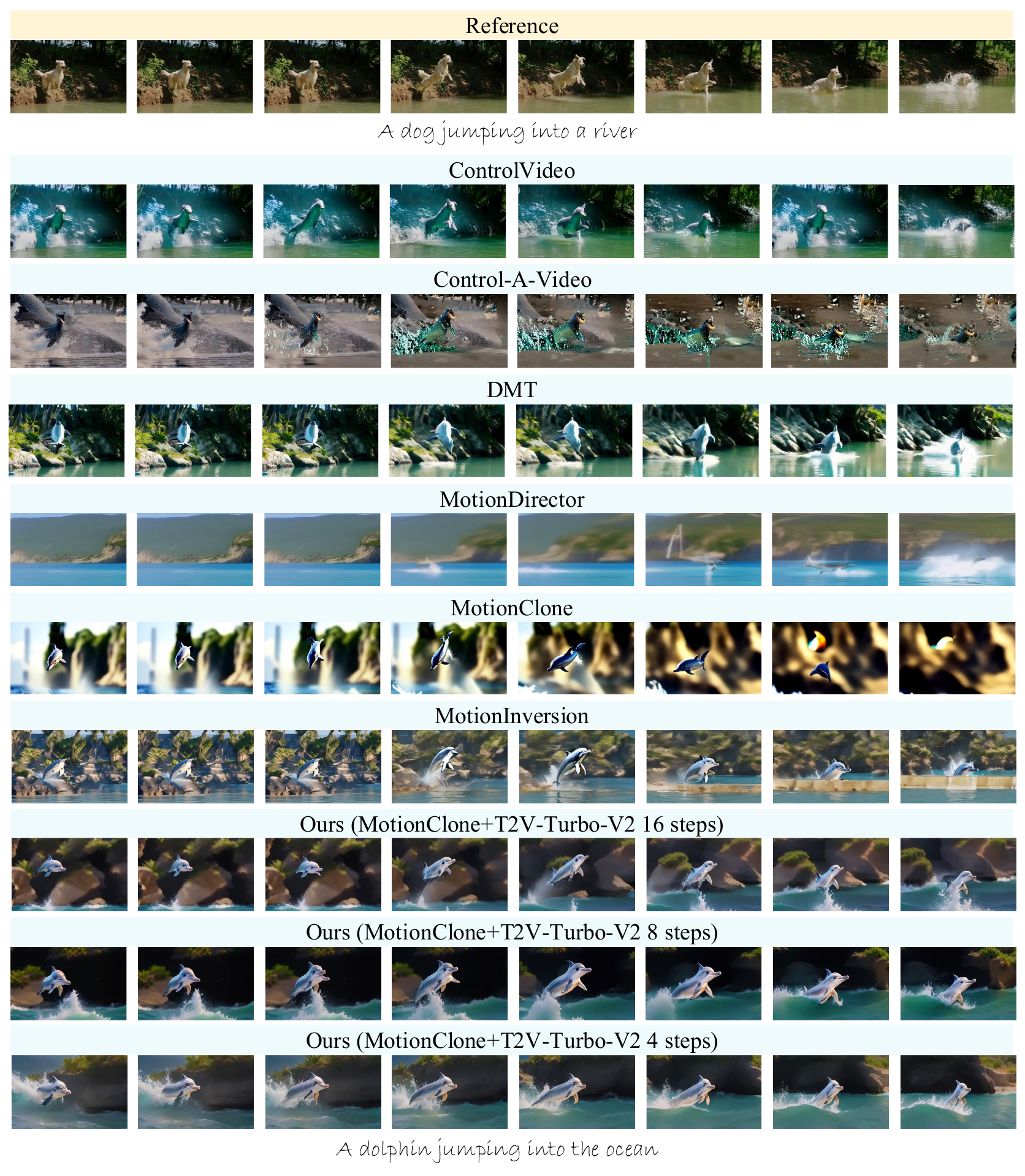}
    \caption{Additional comparison results in hybrid motion customization.}
    \label{fig:c2}
\end{figure*}

\begin{figure*}[t]
    \centering
    \includegraphics[width=1.0\linewidth]{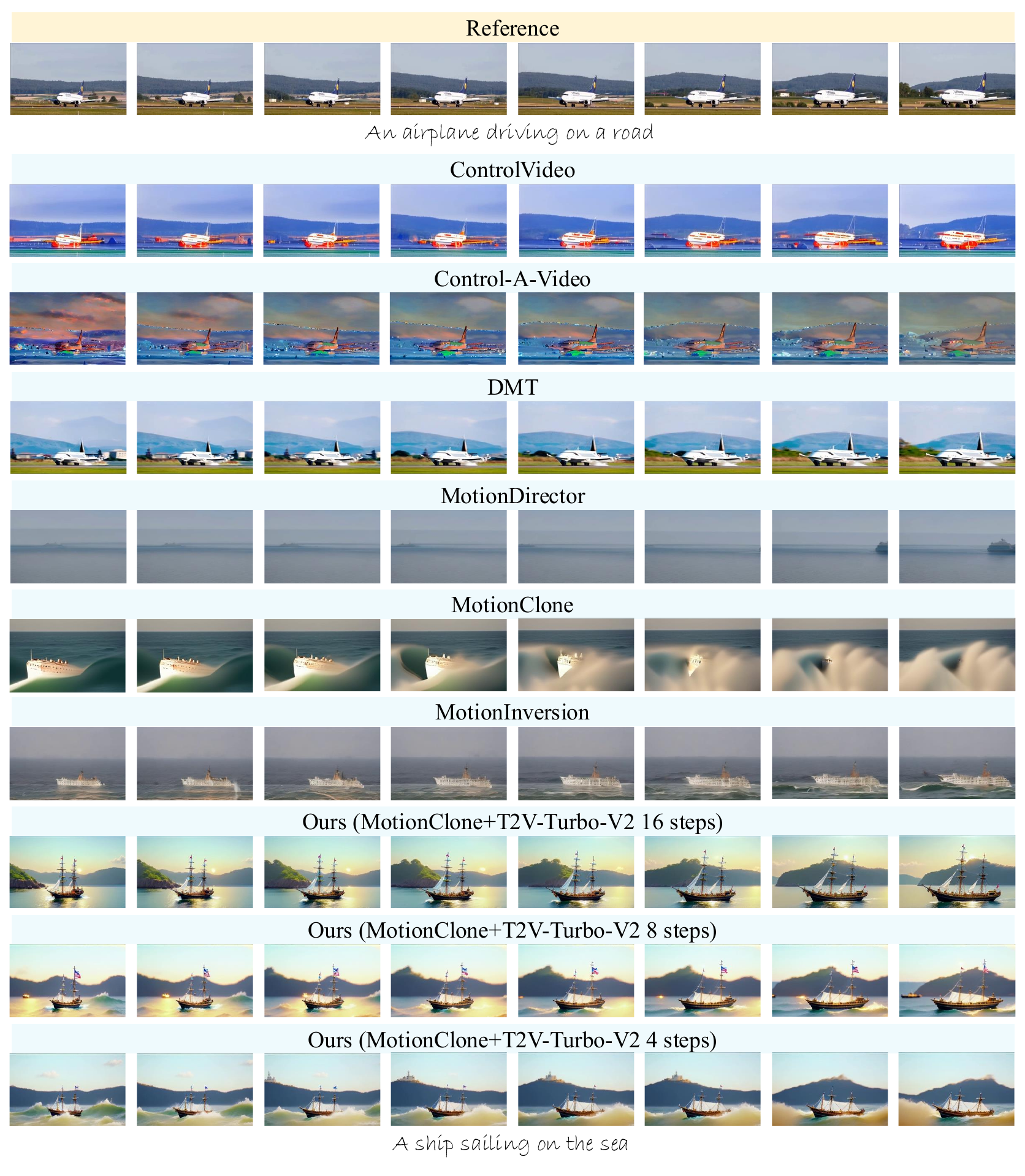}
    \caption{Additional comparison results in hybrid motion customization.}
    \label{fig:c3}
\end{figure*}

\begin{figure*}[t]
    \centering
    \includegraphics[width=1.0\linewidth]{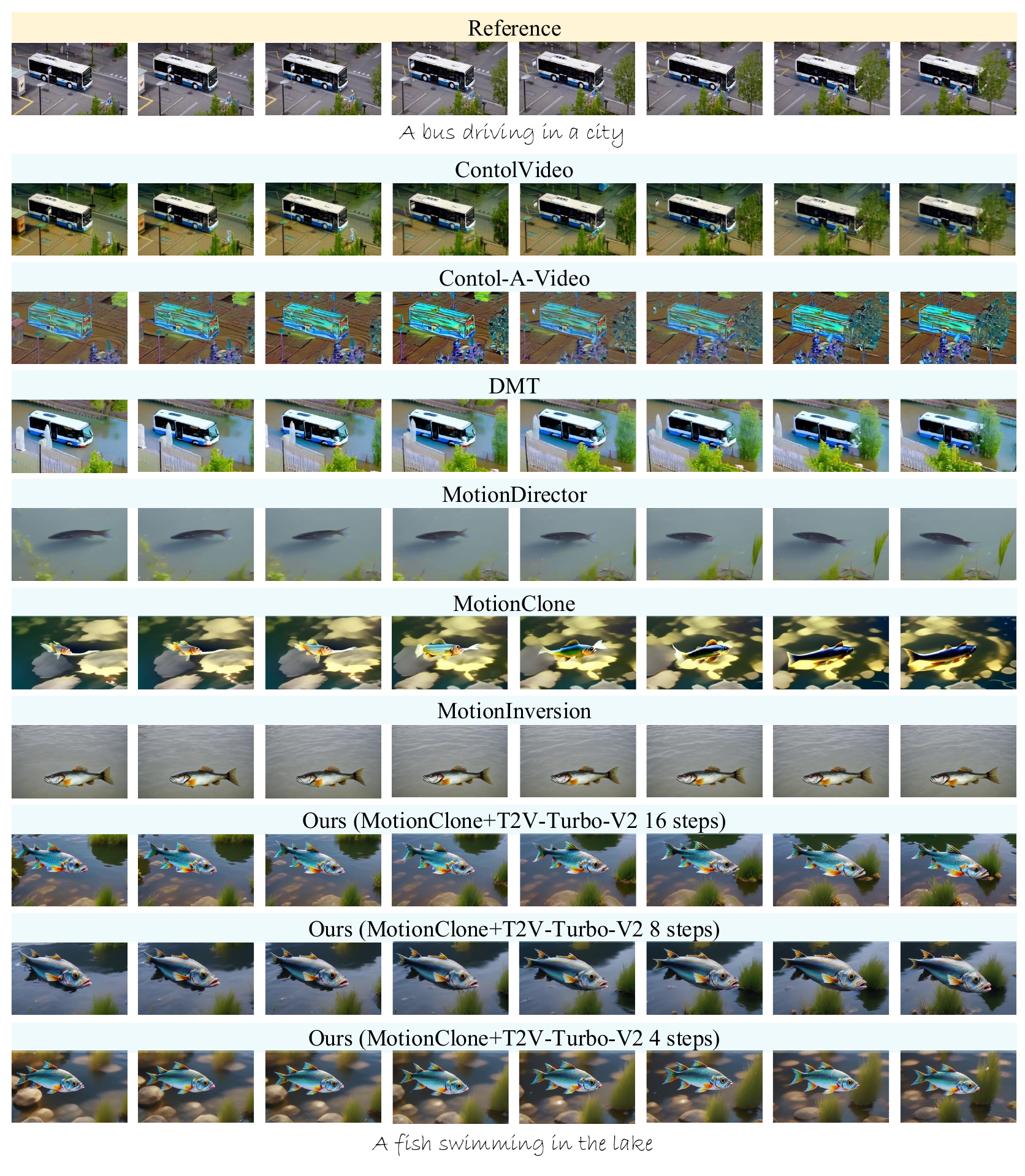}
    \caption{Additional comparison results in hybrid motion customization.}
    \label{fig:c4}
\end{figure*}

\begin{figure*}[t]
    \centering
    \includegraphics[width=1.0\linewidth]{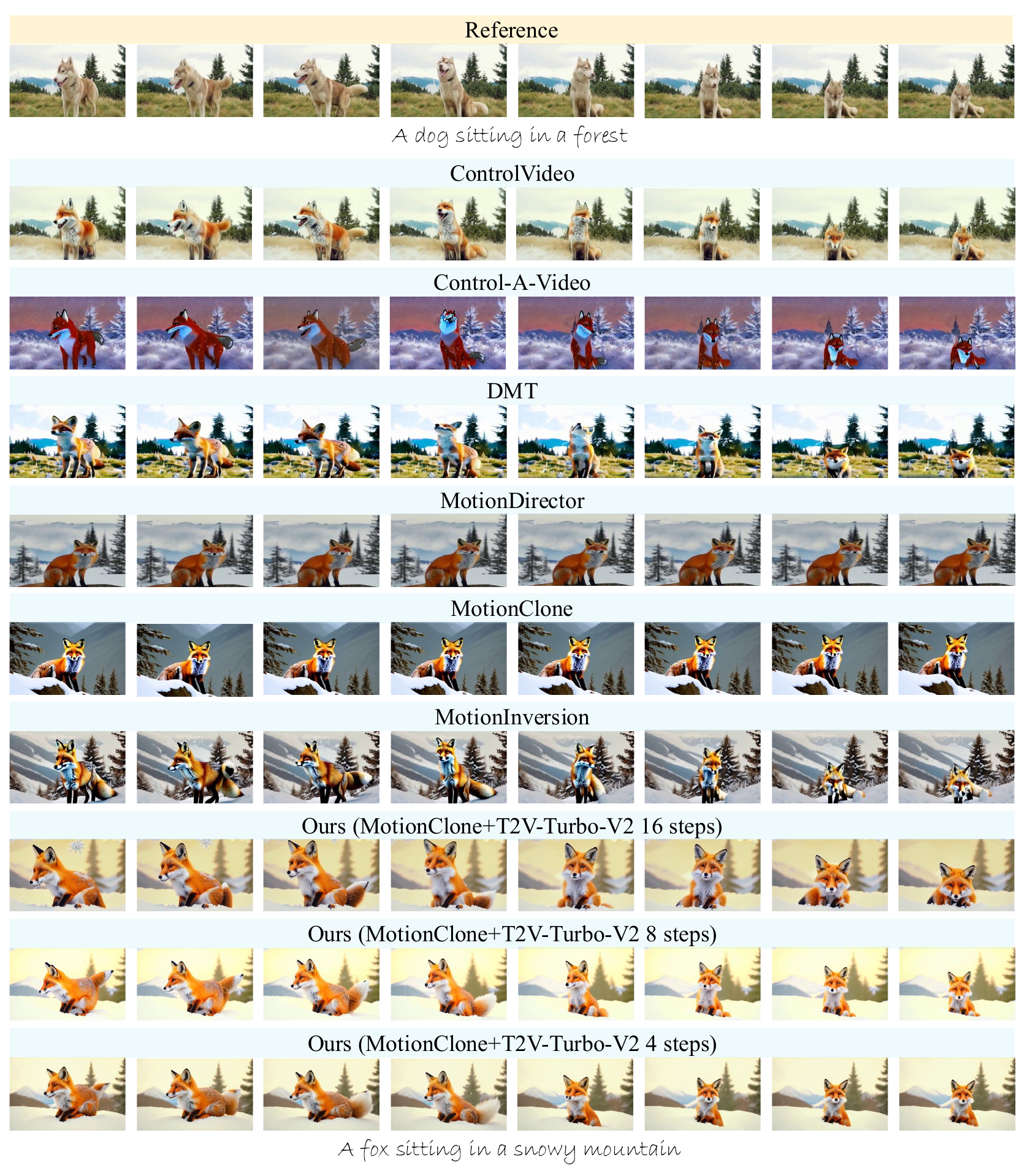}
    \caption{Additional comparison results in object motion customization.}
    \label{fig:c5}
\end{figure*}

\begin{figure*}[t]
    \centering
    \includegraphics[width=1.0\linewidth]{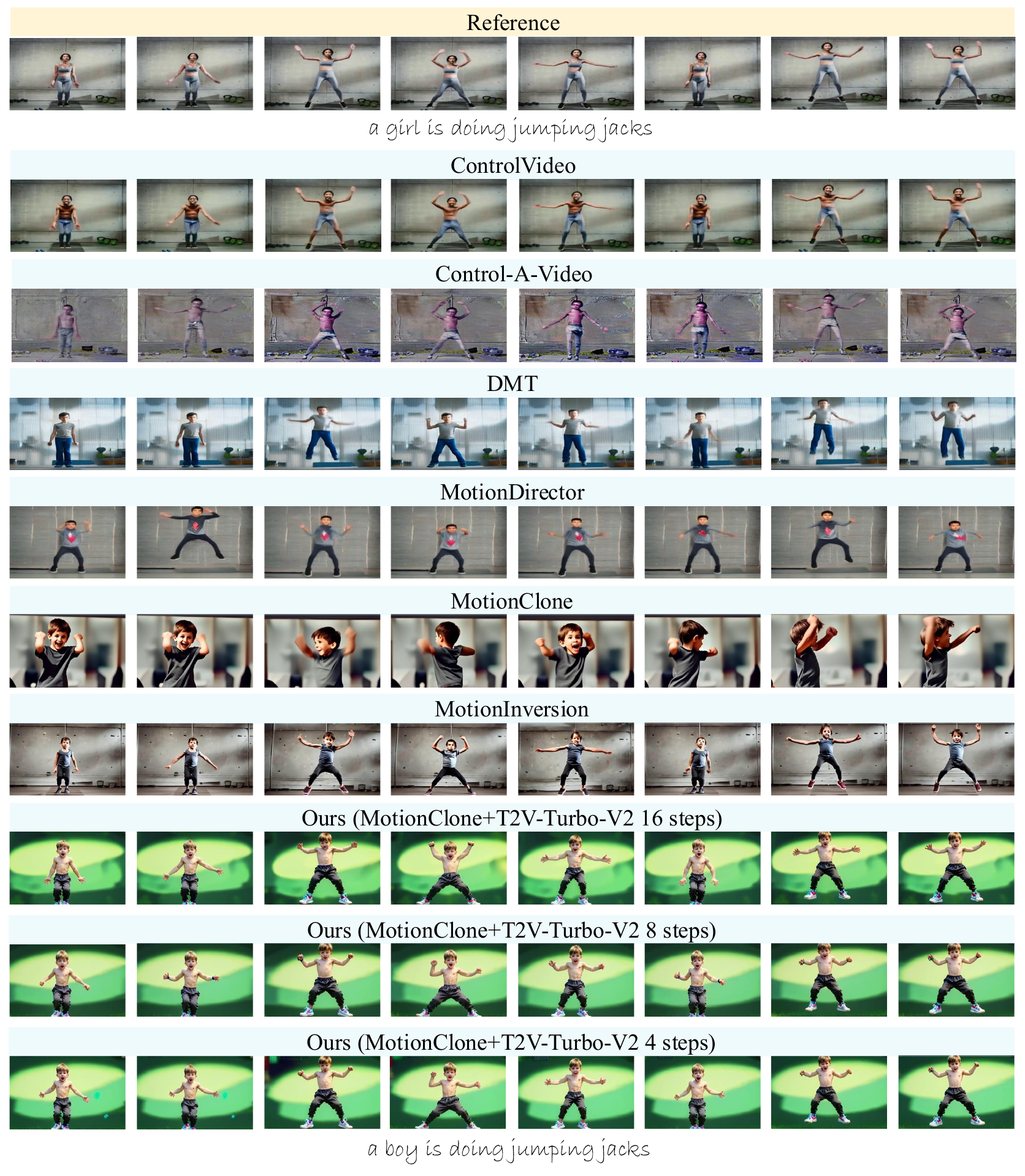}
    \caption{Additional comparison results in object motion customization.}
    \label{fig:c6}
\end{figure*}

\paragraph{More Visual Comparison Results.} In Figure~\ref{fig:compare1} and Figure~\ref{fig:compare2} of the main paper, we provide some exemplar cases of comparing our methods with the other motion customization models including training-based methods (\eg, ControlVideo \cite{ControlVideo}, Control-A-Video \cite{ControlAVideo}, MotionDirector \cite{MotionDirector}, MotionInversion \cite{MotionInversion}) and training-free methods (\eg, MotionClone \cite{MotionClone}, DMT \cite{DMT}). Here, we provide a more comprehensive comparison with more examples of generated videos, covering camera motion (Figure~\ref{fig:c1}), hybrid motion (Figure~\ref{fig:c2}, Figure~\ref{fig:c3}, Figure~\ref{fig:c4}) and object motion (Figure~\ref{fig:c5}, Figure~\ref{fig:c6}). It is observed that our method achieves superior motion fidelity and temporal coherence under the 16-step setting. Remarkably, even when the number of steps is reduced to 8 or 4.
Our method still achieves competitive visual quality, highlighting its efficiency and robustness for fast and high-quality motion customization.

\begin{figure*}[t]
    \centering
    \includegraphics[width=1.0\linewidth]{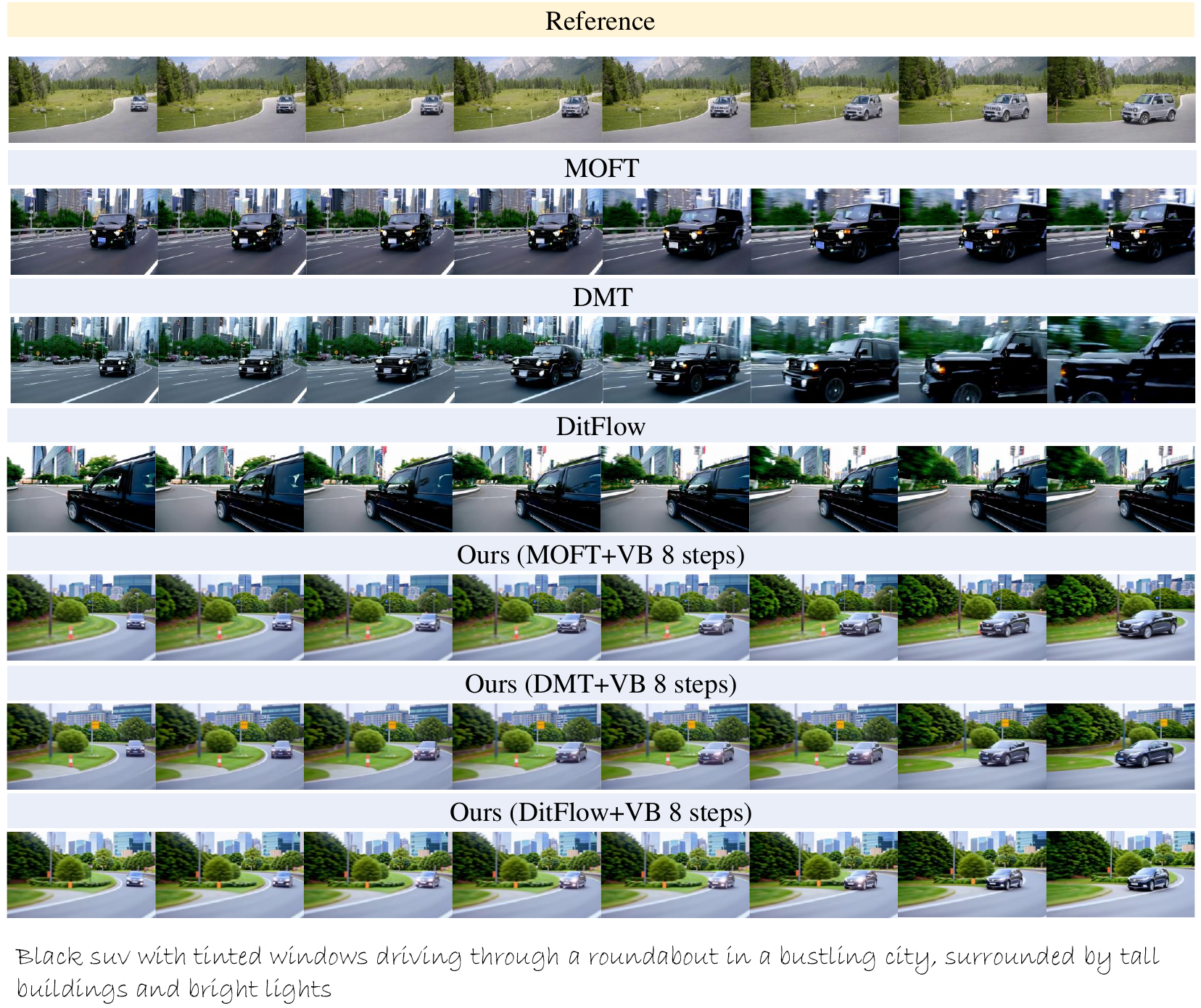}
    \caption{Additional comparison results on Wan2.1-1.3B-based methods.}
    \label{fig:c7}
\end{figure*}

\begin{figure*}[t]
    \centering
    \includegraphics[width=1.0\linewidth]{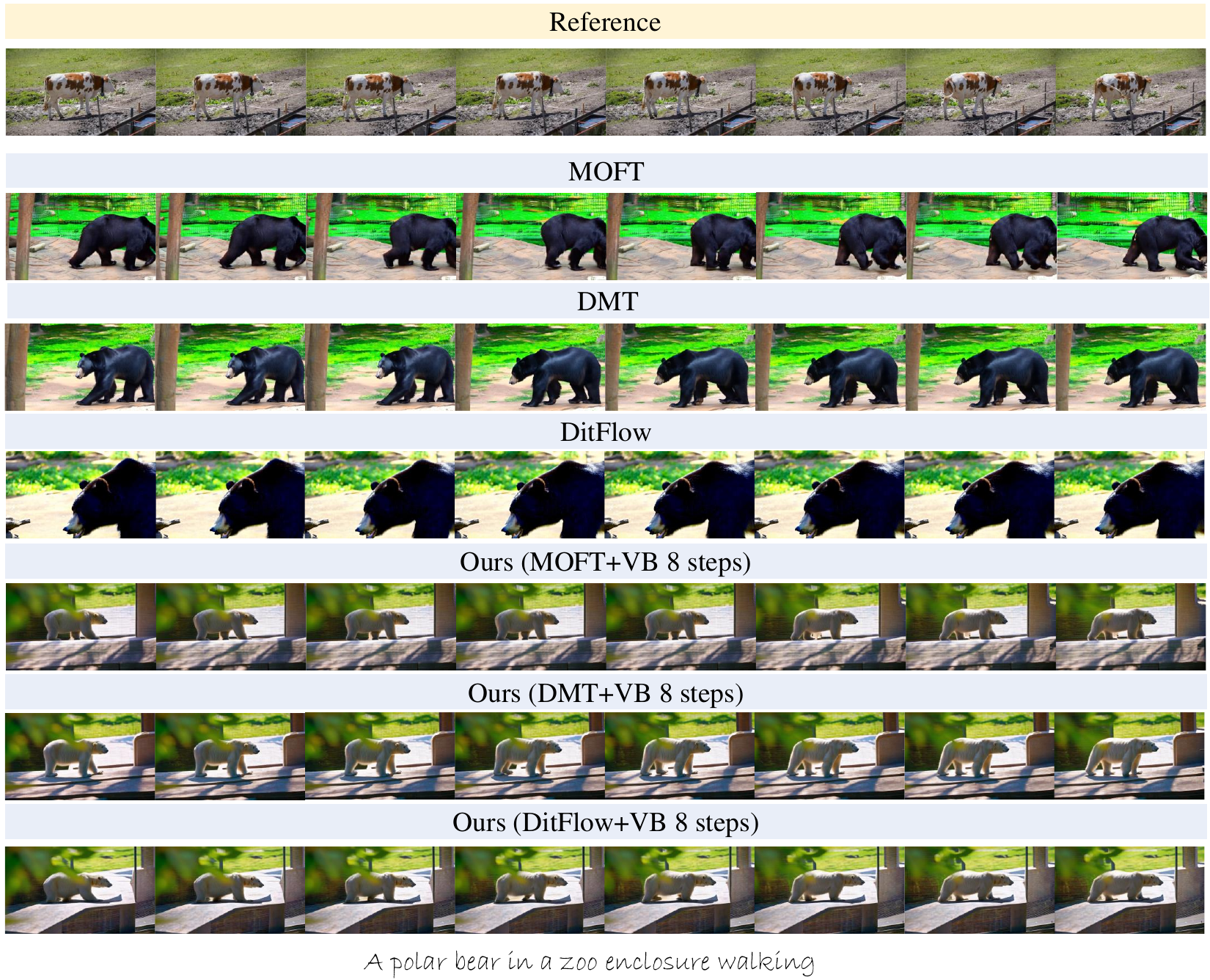}
    \caption{Additional comparison results on Wan2.1-1.3B-based methods.}
    \label{fig:c8}
\end{figure*}

\begin{figure*}[t]
    \centering
    \includegraphics[width=1.0\linewidth]{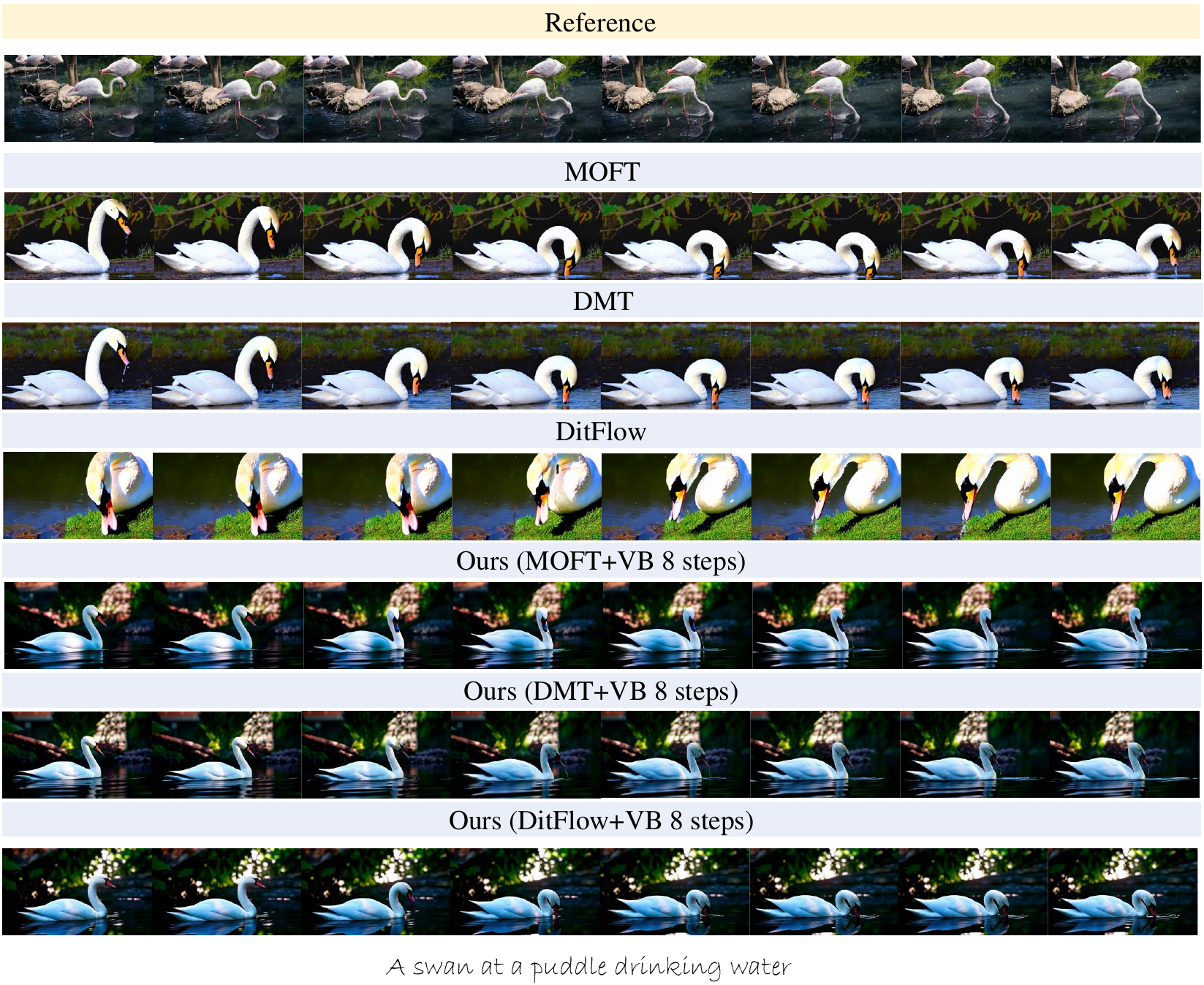}
    \caption{Additional comparison results on Wan2.1-1.3B-based methods.}
    \label{fig:c9}
\end{figure*}

\begin{figure*}[t]
    \centering
    \includegraphics[width=1.0\linewidth]{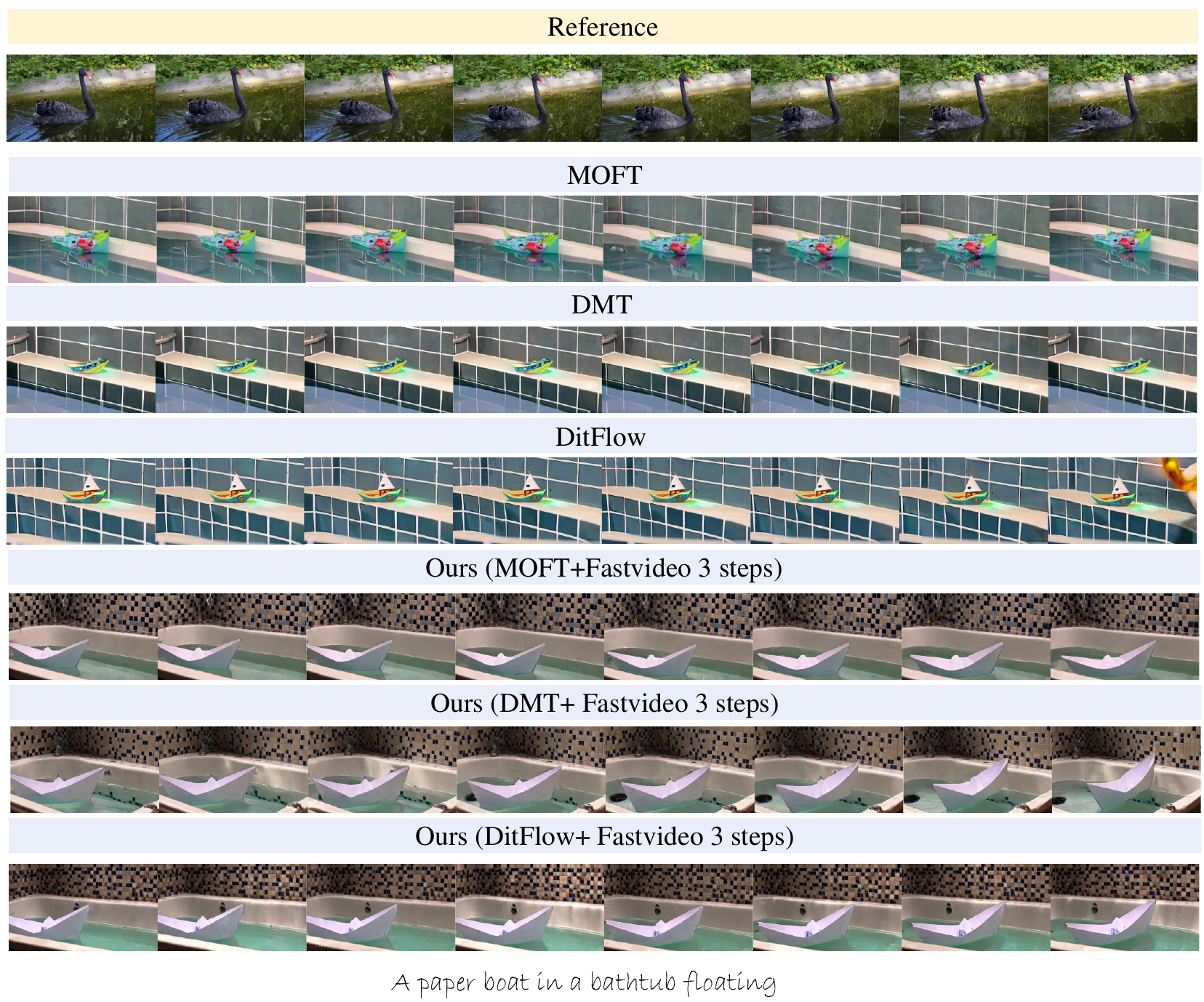}
    \caption{Additional comparison results on Wan2.2-TI2V-5B-based methods.}
    \label{fig:c12}
\end{figure*}

\begin{figure*}[t]
    \centering
    \includegraphics[width=1.0\linewidth]{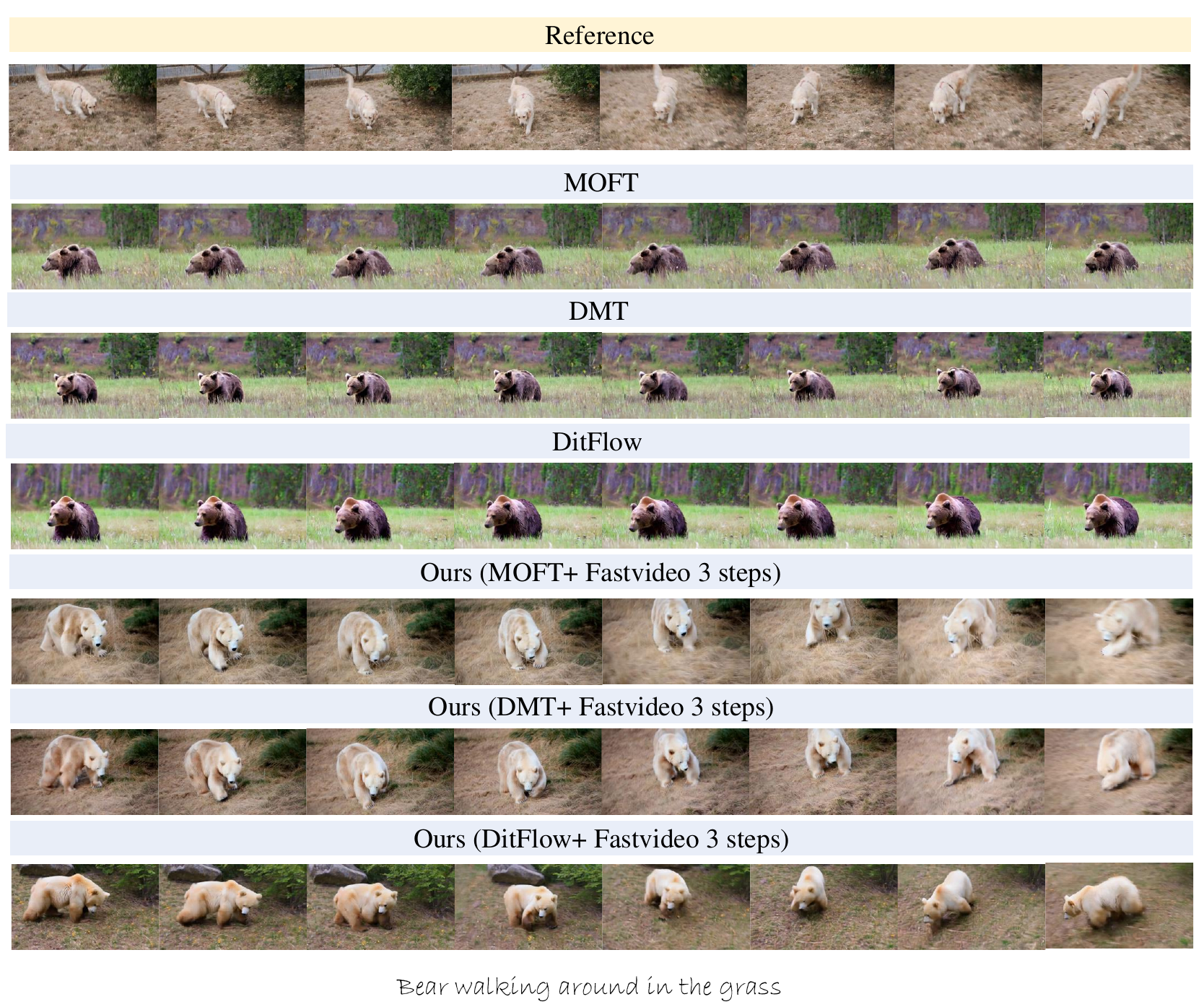}
    \caption{Additional comparison results on Wan2.2-TI2V-5B-based methods.}
    \label{fig:c13}
\end{figure*}

\begin{figure*}[t]
    \centering
    \includegraphics[width=1.0\linewidth]{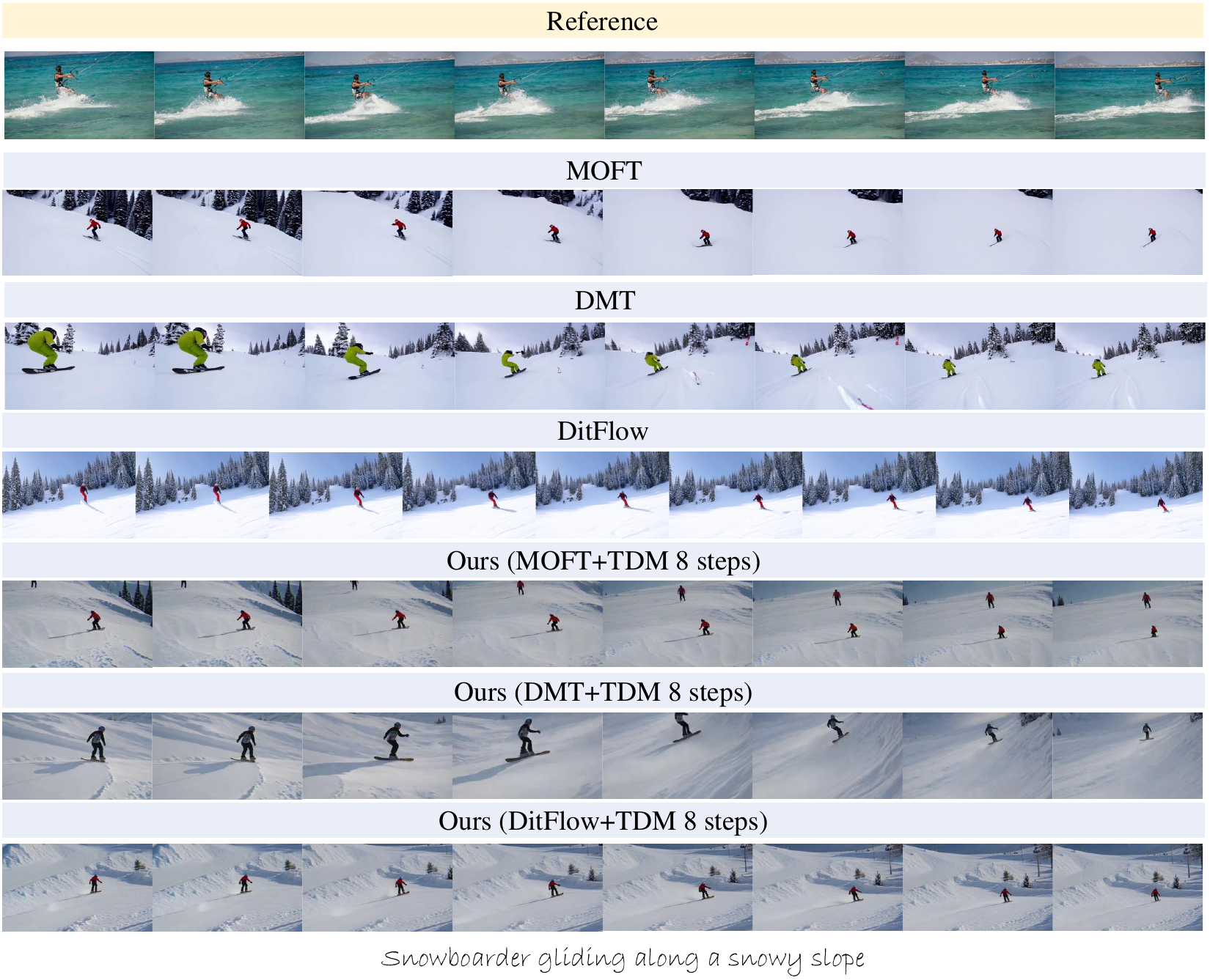}
    \caption{Additional comparison results on CogVideoX-2B-based methods.}
    \label{fig:c10}
\end{figure*}

\begin{figure*}[t]
    \centering
    \includegraphics[width=1.0\linewidth]{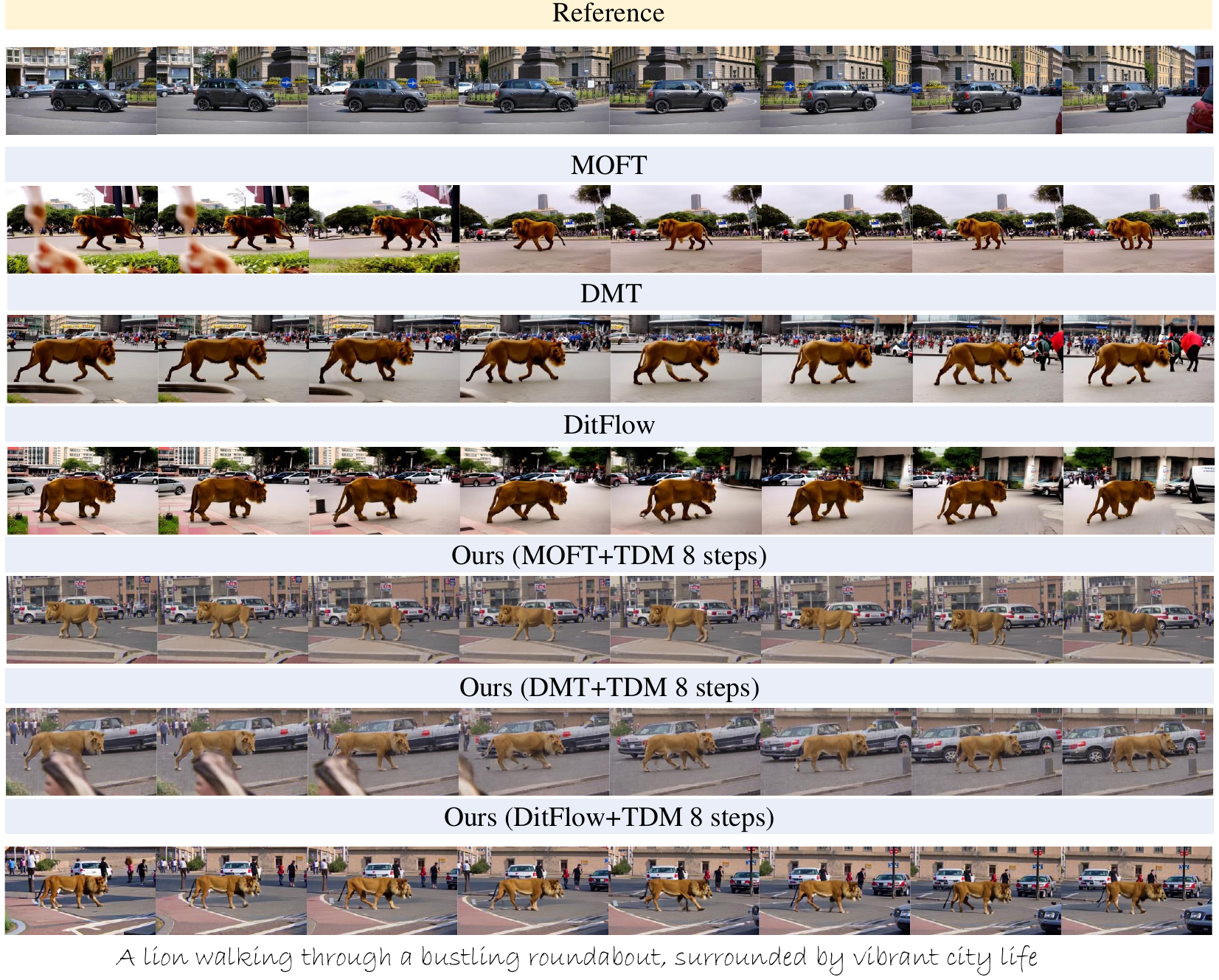}
    \caption{Additional comparison results on CogVideoX-2B-based methods.}
    \label{fig:c11}
\end{figure*}

To further substantiate the scalability of \ourname ~across different DiT architectures, we provide additional comparisons with Wan2.1-1.3B-based, Wan2.2-TI2V-5B-based and CogVideoX-2B-based motion customization methods. Figure~\ref{fig:c7}, Figure~\ref{fig:c8} and Figure~\ref{fig:c9} illustrate the results on VB~\cite{Videoblade} using Wan2.1-1.3B as the teacher model.
Figure~\ref{fig:c12} and Figure~\ref{fig:c13} show the results on Wan2.2-TI2V-5B-Fastvideo using Wan2.2-TI2V-5B as the teacher model.
Figure~\ref{fig:c10} and Figure~\ref{fig:c11} present the results on TDM~\cite{TDM} using CogVideoX-2B as the teacher model. 
These visualizations confirm that our approach effectively generalizes to diverse distilled DiT models, consistently maintaining superior generation quality and motion consistency across varying model scales and architectures.

\begin{figure*}[t]
    \centering
    \includegraphics[width=1.0\linewidth]{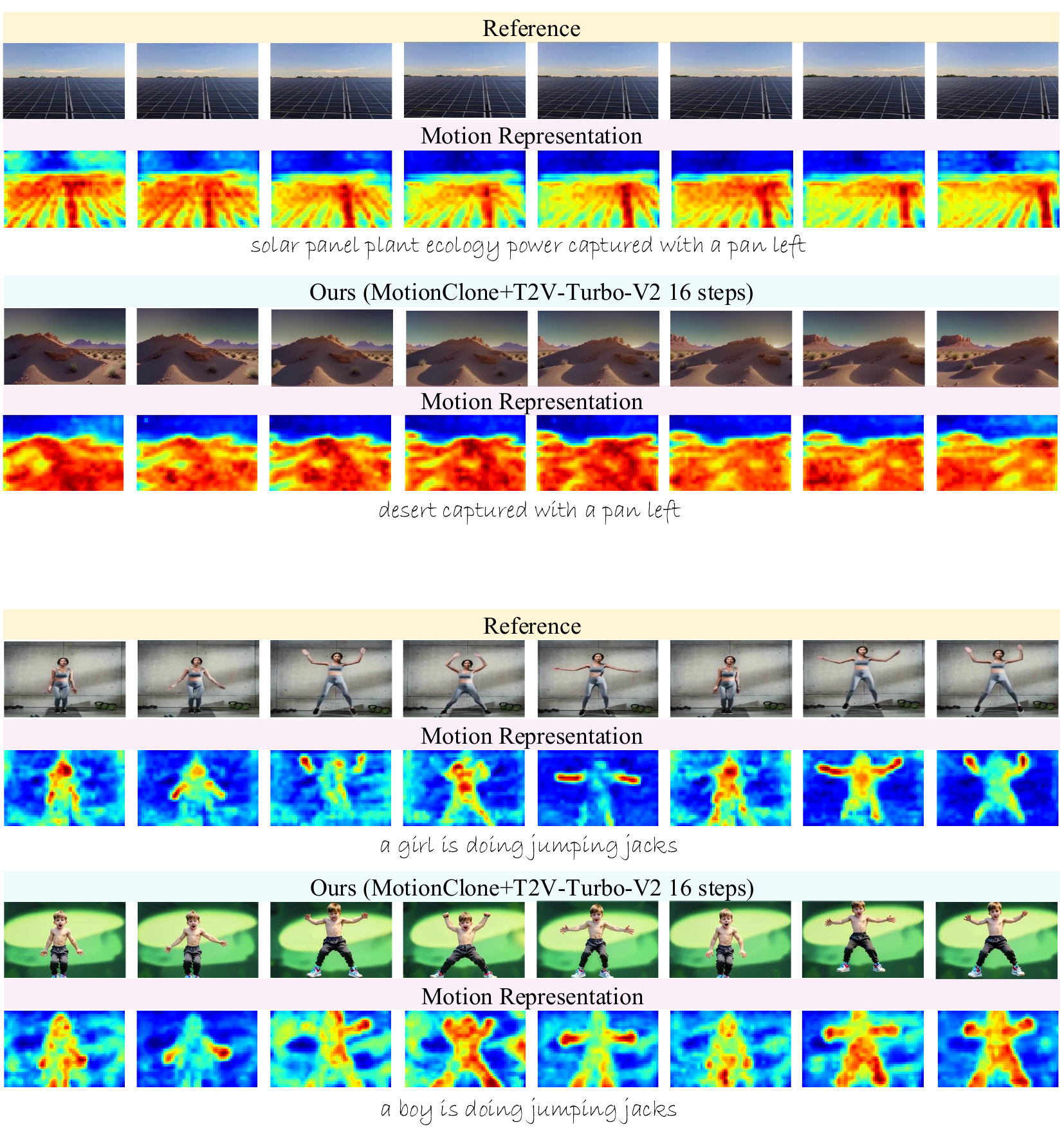}
    \caption{Visualization of motion representation.}
    \label{fig:attn_contra}
\end{figure*}

\paragraph{Visual comparison of attention motion extraction}. Figure~\ref{fig:attn_contra} qualitatively illustrates the effectiveness of our method in aligning motion representations with the reference video, exhibiting strong spatial and temporal consistency.

\begin{table}[t]
    \centering
    \caption{Comparison of peak memory (PM), inference time cost (ITC), and motion fidelity (MF) across different base models.}
    \label{tab:memory_efficiency}
    \begin{tabular}{lcccc}
        \toprule
        Method &Backbone & PM & ITC & MF \\
        \midrule
        MC 16 steps &TurboV2 & 27.00 GB & 7s & 0.833 \\
        Ours (MC 16 steps) &TurboV2 &36.00 GB & 13s & 0.933 \\
        \midrule
        MOFT 8 steps &VB & 10.46 GB & 33s & 0.491 \\
        Ours (MOFT 8 steps) &VB& 12.66 GB & 50s & 0.731 \\
        \midrule
        DMT 8 steps &VB & 10.41 GB & 32s & 0.567 \\
        Ours (DMT 8 steps) &VB& 12.72 GB & 49s & 0.732 \\
        \midrule 
        DiTFlow 8 steps &VB & 15.85 GB & 34s & 0.540 \\
        Ours (DiTFlow 8 steps) &VB & 17.82 GB & 54s & 0.733 \\
        \bottomrule
    \end{tabular}
\end{table}

\begin{figure*}[ht]
    \centering
    \includegraphics[width=1.0\linewidth]{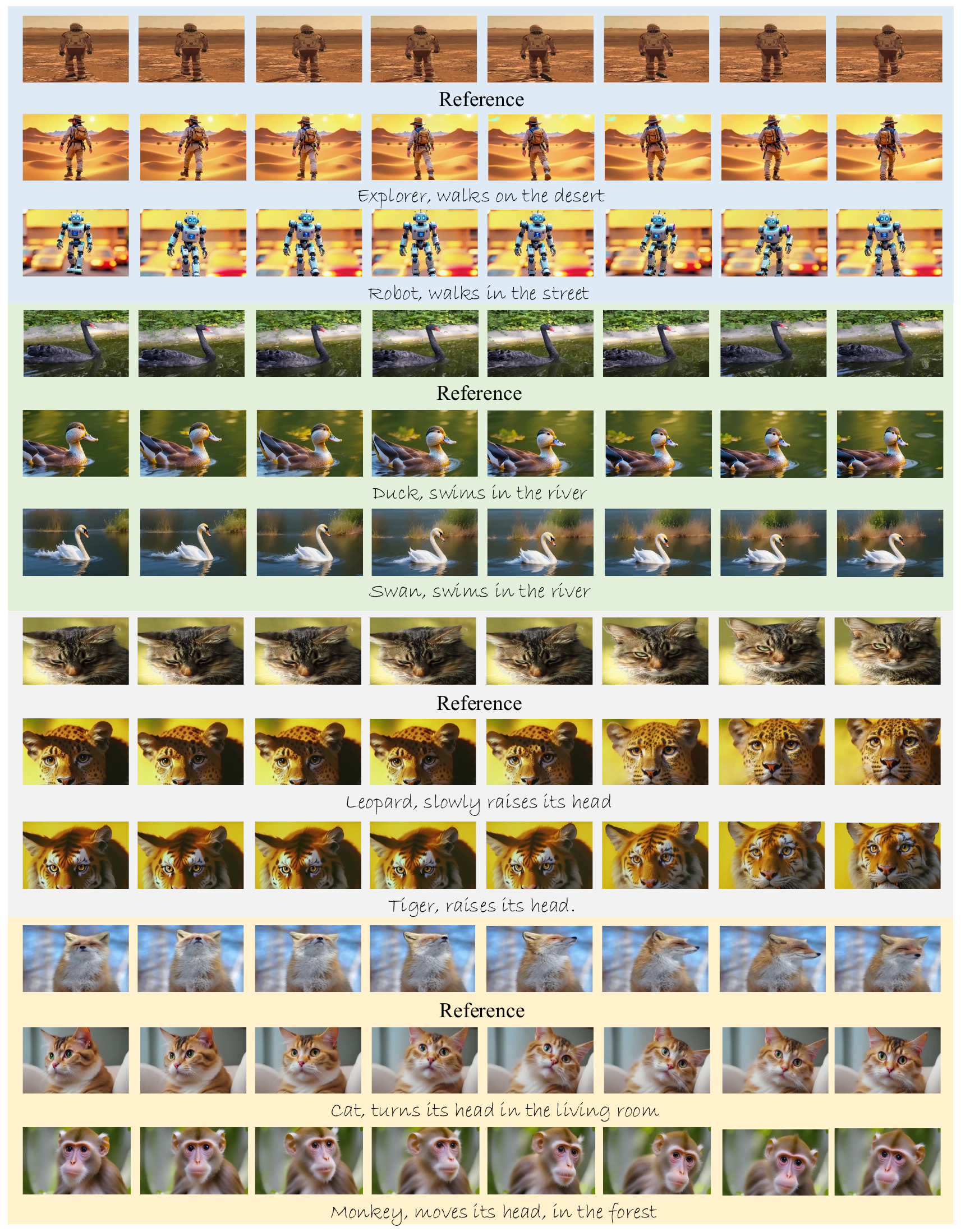}
    \caption{More results of our method (MotionClone+T2V-Turbo-V2) in object motion customization.}
    \label{fig:d1turbo}
\end{figure*}

\begin{figure*}[t]
    \centering
    \includegraphics[width=1.0\linewidth]{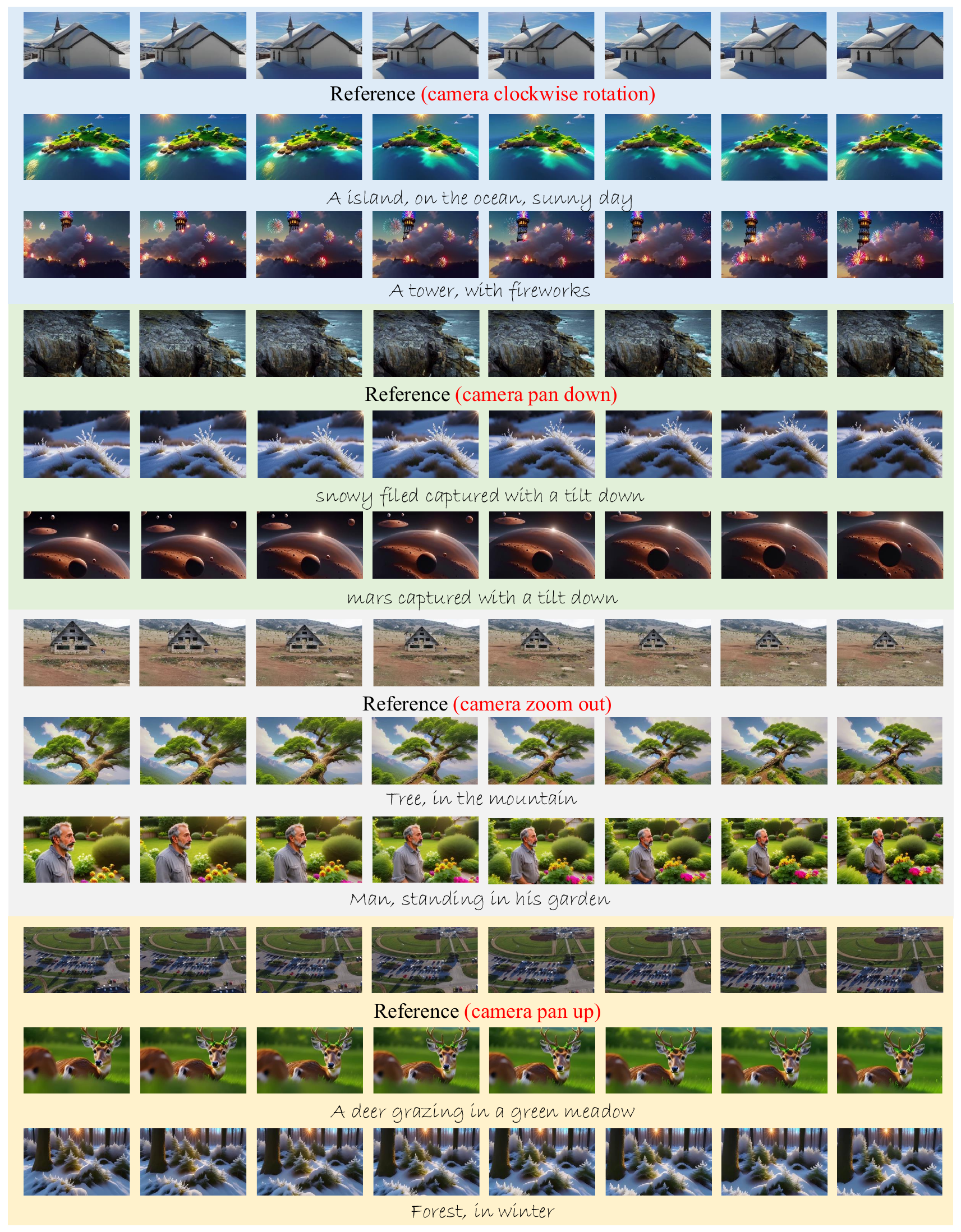}
    \caption{More results of our method (MotionClone+T2V-Turbo-V2) in camera motion customization.}
    \label{fig:d2turbo}
\end{figure*}

\paragraph{Scenario Diversity.} 
For diverse scenarios, \ourname~demonstrates strong adaptability (Figure\ref{fig:d1turbo} and Figure\ref{fig:d2turbo}). For object motion customization, our method successfully generates fine-grained motion for various subjects, such as a duck swimming in a river, a tiger raising its head, and a monkey turning in the forest, preserving semantic consistency while reflecting distinct object behaviors. For camera motion customization, we exhibit faithful reproduction of different camera trajectories including clockwise rotation, zoom out, pan down, pan up, capturing dynamic environmental changes (\eg, fireworks near a tower, a snowy field, or a forest scene). These results validate the robustness of our approach under a wide spectrum of motion types and visual contexts.

\begin{figure*}[t]
    \centering
    \includegraphics[width=1.0\linewidth]{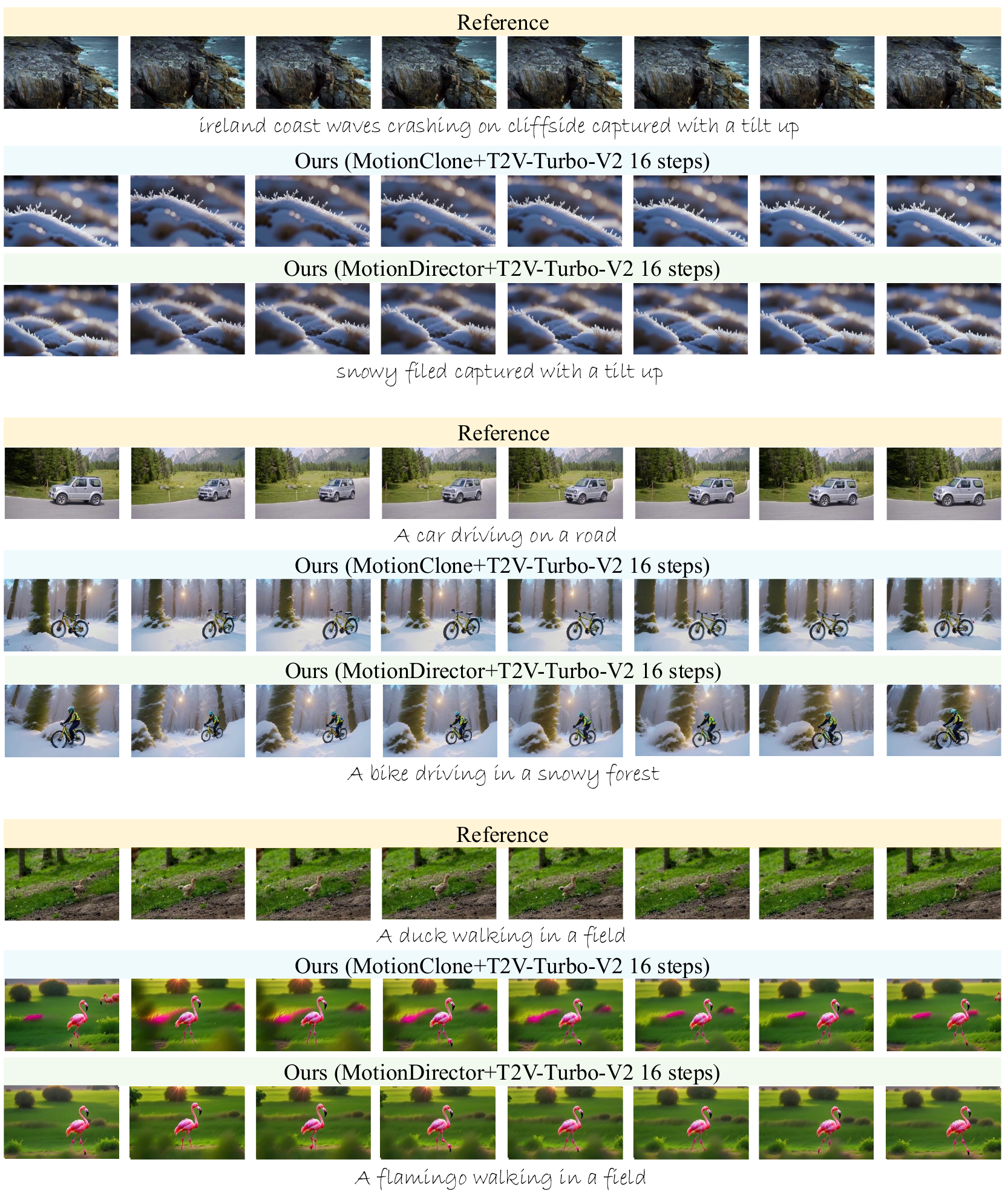}
    \caption{More qualitative results generated by our method under different motion customization strategies.}
    \label{fig:Ours_Generalization}
\end{figure*}

\paragraph{Generalization.} To further verify the flexibility and effectiveness of our method, we evaluate its generalization capability \textit{in terms of motion customization strategies}. Our method is integrated with MotionClone \cite{MotionClone} and MotionDirector \cite{MotionDirector}. As shown in Figure~\ref{fig:Ours_Generalization}, the visual results demonstrate that \ourname~maintains consistent performance with different motion guidance modules.

\subsection{Computational Analysis}
We provide a detailed analysis of GPU memory usage and inference time to evaluate the computational overhead of our method compared to existing approaches on different backbones.

\paragraph{Inference Time and Quality Trade-off.} 
As shown in Table~\ref{tab:memory_efficiency}, \ourname~introduces a strictly bounded and manageable inference overhead. For the TurboV2 backbone, the inference time cost (ITC) increases from 7s to 13s, and for the Video-Blade (VB) backbone, it increases from approximately 33s to 50s. 
This moderate time investment achieves a significant gain in generative quality. 
For instance, across all VB-based baselines, the motion fidelity (MF) surges from an inadequate average of $\sim$0.53 to a robust $\sim$0.73. Considering that full teacher sampling typically requires hundreds of seconds and training-based customization demands hours (as shown in Table~\ref{tab:quantwan}), our adaptive strategy successfully bridges the critical quality gap while strictly preserving the fast generation property of distilled models.

\paragraph{GPU Memory Analysis.} 
The peak memory (PM) on the TurboV2 backbone reaches 36.00 GB, which is entirely acceptable for standard data-center GPUs (e.g., NVIDIA A100). More notably, when applied to the VB backbones, our method demonstrates exceptional memory efficiency. The adaptive intervention mechanism limits the memory overhead to a marginal increase of merely $\sim$2.00 GB over the baselines. With the PM remaining strictly between 12.66 GB and 17.82 GB, \ourname~can be comfortably deployed on consumer-grade GPUs (e.g., RTX 3090/4090 with 24 GB VRAM), making high-fidelity motion customization highly accessible without demanding prohibitive hardware resources.

\section{Derivation of Teacher-Guided Motion Distillation}
\label{app:derivation_tgmd}
We present detailed derivation of the proposed teacher-guided motion distillation. At each student step, we first construct a motion-customized teacher target along the dense teacher trajectory, then transfer this correction to the student via a clean-data-space distillation update. This yields a closed-form interpolation that admits an equivalent formulation in the noise-prediction space. The derivation is presented for diffusion-style students parameterized by noise prediction; for flow-matching-based generators, the same derivation applies by replacing the diffusion-specific clean estimate with the model-native segment endpoint and the noise-prediction correction with a velocity field.

\paragraph{Step-wise setup.}
Consider one student transition from $t_{i+1}$ to $t_i$.
Let
\begin{equation}
s_i
:=
\tilde{z}^{\psi}_{0\leftarrow t_{i+1}}
=
\frac{
z^{\psi}_{t_{i+1}}
-
\sqrt{1-\bar{\alpha}_{t_{i+1}}}\,
\tilde{\epsilon}_{\psi}(z^{\psi}_{t_{i+1}}, c, t_{i+1})
}{
\sqrt{\bar{\alpha}_{t_{i+1}}}
}
\label{eq:der_student_clean}
\end{equation}
denote the student's text-guided clean estimate, where $\tilde{\epsilon}_{\psi}$ is the classifier-free guided prediction defined in the main text.
Starting from $s_i$, we renoise it to an intermediate teacher timestep $t_s \in [t_i,t_{i+1}]$ and perform iterative teacher motion customization.
The resulting motion-customized teacher target is denoted by
\begin{equation}
y_i
:=
\hat{z}^{\theta}_{0\leftarrow t_i}.
\label{eq:der_teacher_target}
\end{equation}

\paragraph{Clean-space distillation update.}
To transfer the teacher correction to the student, we define the step-wise distillation objective
\begin{equation}
\ell_{\mathrm{distill}}
=
\left\|
s_i-\mathrm{sg}[y_i]
\right\|_2^2,
\label{eq:der_loss}
\end{equation}
where $\mathrm{sg}[\cdot]$ denotes stop-gradient.
Directly minimizing Eq.~(\ref{eq:der_loss}) with respect to $z^{\psi}_{t_{i+1}}$ would require the Jacobian
$
\partial \tilde{\epsilon}_{\psi}/\partial z^{\psi}_{t_{i+1}}
$.
We therefore optimize in the clean space and take one gradient step on $s_i$:
\begin{equation}
\hat{z}^{\mathrm{new}}_{0\leftarrow t_{i+1}}
=
s_i-\frac{\lambda}{2}\nabla_{s_i}\ell_{\mathrm{distill}}.
\label{eq:der_gd_step}
\end{equation}
Since
\begin{equation}
\nabla_{s_i}\ell_{\mathrm{distill}}=2(s_i-y_i),
\end{equation}
we obtain
\begin{equation}
\hat{z}^{\mathrm{new}}_{0\leftarrow t_{i+1}}
=
(1-\lambda)s_i+\lambda y_i.
\label{eq:der_interp}
\end{equation}
Finally, the student is updated by
\begin{equation}
z^{\psi}_{t_i}
=
\sqrt{\bar{\alpha}_{t_i}}\,
\hat{z}^{\mathrm{new}}_{0\leftarrow t_{i+1}}
+
\sqrt{1-\bar{\alpha}_{t_i}}\,\epsilon_{t_i}.
\label{eq:der_student_step}
\end{equation}

\paragraph{Equivalent proximal form.}
Eq.~(\ref{eq:der_interp}) is also the solution of the proximal problem
\begin{equation}
\hat{z}^{\mathrm{new}}_{0\leftarrow t_{i+1}}
=
\arg\min_x
\frac{1}{2}\|x-s_i\|_2^2
+
\frac{\mu}{2}\|x-y_i\|_2^2.
\label{eq:der_prox}
\end{equation}
Solving the first-order optimality condition gives
\begin{equation}
\hat{z}^{\mathrm{new}}_{0\leftarrow t_{i+1}}
=
\frac{1}{1+\mu}s_i+\frac{\mu}{1+\mu}y_i.
\label{eq:der_prox_closed}
\end{equation}
By setting
\begin{equation}
\lambda=\frac{\mu}{1+\mu},
\qquad
\mu=\frac{\lambda}{1-\lambda},
\label{eq:der_lambda_mu}
\end{equation}
we recover Eq.~(\ref{eq:der_interp}).
Hence, the update can be viewed as balancing fidelity to the student trajectory and attraction to the teacher-corrected target.

\paragraph{Noise-prediction form.}
The interpolation in Eq.~(\ref{eq:der_interp}) can be further rewritten in the noise-prediction space.
Define the teacher-induced residual at the current student state as
\begin{equation}
\epsilon_i^{\mathrm{echo}}
:=
\frac{
z^{\psi}_{t_{i+1}}
-
\sqrt{\bar{\alpha}_{t_{i+1}}}\,y_i
}{
\sqrt{1-\bar{\alpha}_{t_{i+1}}}
}.
\label{eq:der_teacher_residual}
\end{equation}
Then Eq.~(\ref{eq:der_interp}) is equivalent to
\begin{equation}
\hat{z}^{\mathrm{new}}_{0\leftarrow t_{i+1}}
=
\frac{
z^{\psi}_{t_{i+1}}
-
\sqrt{1-\bar{\alpha}_{t_{i+1}}}\,
\tilde{\epsilon}_i
}{
\sqrt{\bar{\alpha}_{t_{i+1}}}
},
\label{eq:der_reparam_clean}
\end{equation}
where
\begin{equation}
\tilde{\epsilon}_i
=
(1-\lambda)\tilde{\epsilon}_{\psi}(z^{\psi}_{t_{i+1}}, c, t_{i+1})
+
\lambda \epsilon_i^{\mathrm{echo}}.
\label{eq:der_guided_eps}
\end{equation}
Equivalently,
\begin{equation}
\tilde{\epsilon}_i
=
\tilde{\epsilon}_{\psi}(z^{\psi}_{t_{i+1}}, c, t_{i+1})
-
\lambda
\Bigl(
\tilde{\epsilon}_{\psi}(z^{\psi}_{t_{i+1}}, c, t_{i+1})
-
\epsilon_i^{\mathrm{echo}}
\Bigr).
\label{eq:der_guidance_form}
\end{equation}
This shows that teacher-guided distillation can be interpreted as a correction in the noise-prediction space.

\begin{figure}[t]
    \centering
    \includegraphics[width=1.0\linewidth]{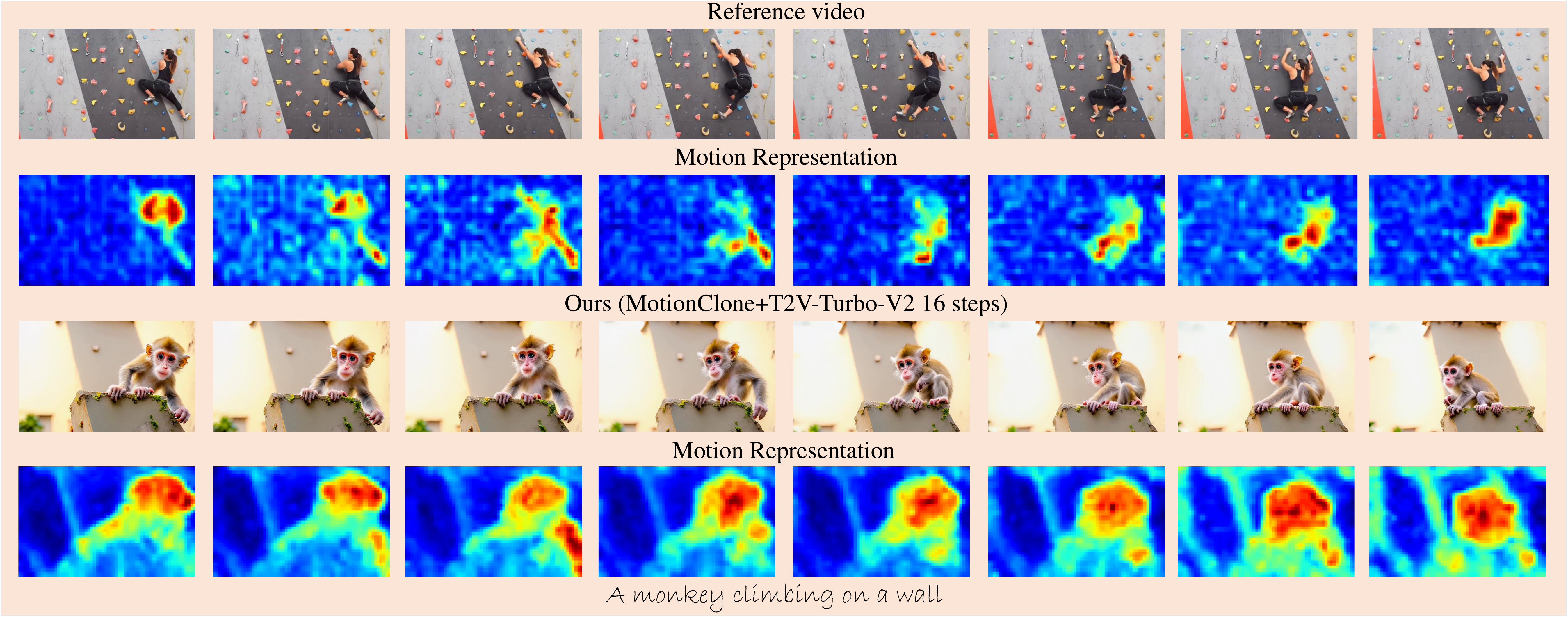}
    \caption{\textbf{Failure case.} Reference video ``A woman climbing on a wall" (rows 1-2: frames and motion representation) versus generated ``A monkey climbing on a wall" (rows 3-4: frames and motion representation). Note the diffuse and over-activated motion features in the generated result (Row 4) compared to the precise climb activations in the reference (Row 2).}
    \label{fig:limitations}
\end{figure}

\section{Limitations and Future Work} 
\label{app:limit}

While \ourname~demonstrates strong performance across a wide range of scenarios, it has certain limitations that warrant discussion. First, our method may encounter challenges when the target object-motion combination lies outside the training distribution of the base text-to-video (T2V) model, or when there is a significant structural discrepancy between the source and target subjects. 

For instance, in the object task of transforming ``A woman climbing on a wall" to ``A monkey climbing on a wall," as shown in Figure~\ref{fig:limitations}, we observe a noticeable degradation in motion precision. Comparing the motion representations, the reference video (Row 2) exhibits clear, localized activation patterns corresponding to specific limb movements. In contrast, the generated video (Row 4) displays over-activated motion features. This indicates that the model struggles to disentangle and spatially align the fine-grained climbing motion of a human with the monkey's body structure, resulting in a loss of detailed motion control. Similar failure modes have been reported in related training-free methods~\cite{MotionClone,MotionInversion,DitFlow}.
We attribute this breakdown to two factors: (1) the generative priors of the T2V model cannot faithfully encode such fine-grained motion when the object-motion pairing is underrepresented in the training data, and (2) since \ourname~operates primarily in the latent space, attention mechanisms struggle to represent local subtle motions, such as the nuanced movement of fingers or distinct limb coordination during climbing actions. 

In future work, we plan to investigate several directions to address these limitations. We will explore methods to improve the representation of fine-grained local motions, possibly by incorporating hierarchical attention mechanisms or explicit motion modeling in pixel space. Moreover, we will explore unpaired teacher models with unshared latent spaces, a challenging open problem in diffusion model distillation, and investigate self-assessment mechanisms enabling more efficient and selective speculative guidance during inference.

\section{Ethical and Social Impacts}
\label{app:ethical}

The development of \ourname, a training-free test-time distillation framework for motion customization, is a double-edged sword. On one hand, our method enables efficient and flexible motion transfer without any training, which opens up new possibilities for creative industries, education, and accessibility by allowing users to generate personalized videos with minimal resources.

On the other hand, the powerful capabilities of \ourname~raise important concerns regarding potential misuse. Our method relies on the pre-trained text-to-video diffusion models, which are trained on large-scale datasets sourced from the internet or generated content. These datasets may contain biased, inappropriate, unsafe or copyrighted material, potentially leading to the inadvertent propagation of societal stereotypes or unauthorized replication of proprietary content. Additionally, the ability to generate highly realistic motion-aligned videos with minimal effort increases the risk of misuse, such as fabricating misleading media or deepfakes. To mitigate these risks, we advocate for the implementation of ethical safeguards, including automated content filtering, watermark-based provenance tracking, and clearly defined usage policies that prioritize transparency, accountability, and user consent. 

To promote transparency in AI-generated content and foster further research in video generation, the source code will be made publicly available upon acceptance of the paper. We hope that open access to this technology will catalyze collaborative efforts toward safer and more inclusive generative systems.

\end{document}